\documentclass[runningheads]{llncs}

\usepackage{eccv}

\usepackage{eccvabbrv}

\usepackage{graphicx}
\usepackage{booktabs}
\usepackage{multirow}
\usepackage{adjustbox}
\usepackage{footnote}
\usepackage{enumitem}
\usepackage[accsupp]{axessibility}  

\usepackage[pagebackref,breaklinks,colorlinks,citecolor=eccvblue]{hyperref}

\usepackage{orcidlink}

\makeatletter
\def\blfootnote#1{\xdef\@thefnmark{}\@footnotetext{\scriptsize #1}}
\makeatother

\begin{document}

\title{HiDiffusion:  Unlocking Higher-Resolution Creativity and Efficiency in Pretrained Diffusion Models} 

\titlerunning{HiDiffusion}

\author{Shen Zhang \and
Zhaowei Chen \and
Zhenyu Zhao \and
Yuhao Chen \and
Yao Tang \and 
\quad Jiajun Liang\inst{\dagger}}

\authorrunning{Shen Zhang et al.}

\institute{MEGVII Technology \\
\email{\{zhangshen1915, chaowechan, yhao.chen0617\}@gmail.com \\
\{zhaozhenyu, tangyao02, liangjiajun\}@megvii.com} \\
\vspace{0.5em}
Project Page: \url{https://hidiffusion.github.io/}
}

\maketitle

\blfootnote{$^\dagger$Corresponding author}

\begin{abstract}

Diffusion models have become a mainstream approach for high-resolution image synthesis. However,  directly generating \textbf{higher-resolution} images from pretrained diffusion models will encounter unreasonable object duplication and exponentially increase the generation time. In this paper, we discover that object duplication arises from feature duplication in the deep blocks of the U-Net.  Concurrently, We pinpoint the extended generation times to self-attention redundancy in U-Net's top blocks. To address these issues, we propose a tuning-free higher-resolution framework named HiDiffusion. Specifically, HiDiffusion contains Resolution-Aware U-Net~(RAU-Net) that dynamically adjusts the feature map size to resolve object duplication and engages Modified Shifted Window Multi-head Self-Attention~(MSW-MSA) that utilizes optimized window attention to reduce computations. 
we can integrate HiDiffusion into various pretrained diffusion models to scale image generation resolutions even to 4096×4096 at 1.5-6× the inference speed of previous methods. Extensive experiments demonstrate that our approach can address object duplication and heavy computation issues, achieving state-of-the-art performance on higher-resolution image synthesis tasks.

\end{abstract}

\section{Introduction}
\label{sec:intro}

\begin{figure}[tb]
  \centering
  \includegraphics[width=\textwidth]{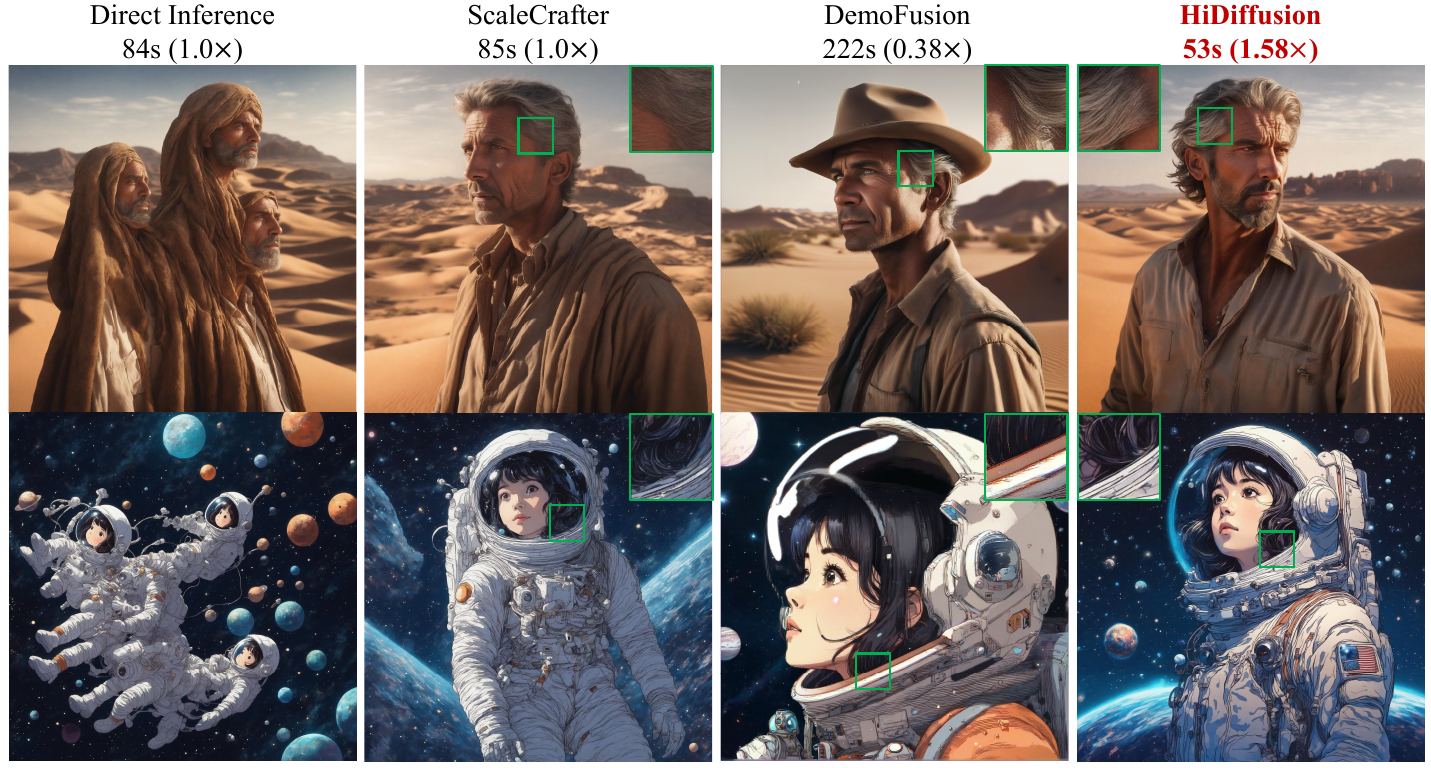}
  \caption{2048$\times$2048 resolution images based on SDXL~\cite{podell2023sdxl}. The first line of the text indicates the generation methods, while the second line indicates the cost time and inference speed relative to direct inference. Our Hidiffusion can generate reasonable and realistic high-resolution images with high efficiency. Compared to previous methods, ours exhibits richer fine-grained details and is 1.58$\times$ faster than Scalerafter~\cite{he2023scalecrafter}, 4.18$\times$ faster than DemoFusion~\cite{du2023demofusion}. Best viewed when zoomed in.
  }
  \label{fig:call_back}
  \vspace{-2em}
\end{figure}

Generative model has witnessed an explosion of diffusion models of growing capability and applications~\cite{song2019generative,ho2020denoising,song2020score,song2020denoising,rombach2022high}. Being trained on a large volume of images~(Laion 5B~\cite{laion}), Stable Diffusion~(SD)~\cite{rombach2022high,podell2023sdxl} can generate fixed-size~(e.g. 512$\times$512 for SD 1.5~\cite{rombach2022high}) high-quality images given text or other kinds of prompts. However, it is limited to synthesizing images with higher resolutions~(e.g. 2048$\times$2048). The limitation has two perspectives:  \textbf{(\romannumeral1) Feasibility}. Diffusion models lack scalability in higher-resolution image generation. As illustrated in the 
 Direct Inference column of ~\cref{fig:call_back}, when directly inferencing to generate 2048$\times$2048 resolution images for SDXL~\cite{podell2023sdxl} that being trained on 1024$\times$1024 resolution, the generated images exhibit unreasonable object duplication and inexplicable object overlaps. 
\textbf{(\romannumeral2) Efficiency.} As resolution increases, the time cost becomes more and more unacceptable. For example, SD 1.5 can generate a 512$\times$512 resolution image in only 3s, whereas it takes 165s to generate a 2048$\times$2048 image on an NVIDIA V100 with 50 DDIM steps. The low efficiency of diffusion models in higher-resolution synthesis makes it impractical for real-world applications. We ask: \textbf{Can Stable Diffusion efficiently synthesize images with resolution beyond the training image sizes?}

  Existing methods answer feasibility question from three perspectives: 
  (\romannumeral1) Collecting enough higher-resolution images to retrain a diffusion model for higher-resolution synthesis~\cite{podell2023sdxl}.
  (\romannumeral2) Leveraging additional super-resolution models~\cite{rombach2022high,zheng2023any} to upscale low-resolution images.   (\romannumeral3) Modifying the operation~\cite{jin2023training} or architecture~\cite{he2023scalecrafter} of U-Net, or creating a new synthesis schedule~\cite{du2023demofusion} for higher-resolution synthesis in a tuning-free way.  While those perspectives can mitigate object duplication, the first two require large-scale high-resolution datasets and the training process is costly. The tuning-free methods can leverage the power of the pretrained diffusion model, but they suffer from insufficient image details or low inference efficiency, as shown in~\cref{fig:call_back}.

In this paper, we aim to resolve object duplication and generate higher-resolution images with fine details in a tuning-free way. 
Different from previous methods, we explore a new perspective by investigating the feature map in the U-Net. Our observation reveals that the generated image is highly correlated with the feature map of deep blocks in structures and feature duplication happens in the deep Blocks. The highly duplicated features guide the synthesis direction, resulting in object duplication. We propose a simple yet effective method called Resolution-Aware U-Net (RAU-Net). RAU-Net involves Resolution-Aware Downsampler (RAD) and Resolution-Aware Upsampler (RAU) to align the feature map size with the deep block of U-Net. 
In contrast to ScaleCrafter~\cite{he2023scalecrafter}, which attributes object duplication to the limited receptive field of convolutions and requires determining the parameters of \textit{each} convolution in U-Net, we discover feature duplication and propose a more concise solution RAU-Net that modifies the parameters of only \textit{two} convolutions.
Therefore, RAU-Net can be more readily integrated into various diffusion models. The slight modifications can also better preserve the capabilities of the pretrained model, and consequently retain more fine-grained image details. To further improve the higher-resolution image quality, we propose a Switch Threshold to boost the fine details of generated higher-resolution images. Our proposed approach requires no further fine-tuning and can be seamlessly integrated into diffusion models.

Besides higher-resolution feasibility, efficiency is another important concern. While numerous works focus on reducing the sampling step~\cite{lu2022dpm,song2020denoising,salimans2022progressive,meng2023distillation,lu2022dpm++}, few studies investigate the acceleration of diffusion U-Net~\cite{bolya2023token,ma2023deepcache}. These acceleration methods enhance the inference efficiency but also compromise the generated image quality. In this paper, we unearth that the dominant time-consuming global self-attention in the top blocks exhibits surprising locality. Inspired by this observation, we propose Modified Shifted Window Multi-head Self-Attention (MSW-MSA) and replace the global self-attention with it in higher-resolution synthesis.  This substitution needs no further fine-tuning. Compared to the previous local attention method~\cite{liu2021swin}, our method uses large window attention and shifts windows across timestep to accommodate diffusion models. Empirically, MSW-MSA achieves significant acceleration without compromising image quality.

We combine RAU-Net and MSW-MSA into a unified tuning-free framework for higher-resolution image generation, dubbed HiDiffusion. We conduct qualitative
and quantitative experiments to validate the effectiveness of our method. Specifically, HiDiffusion can scale the resolution of SD 1.5~\cite{rombach2022high} and SD 2.1~\cite{rombach2022high} from 512$\times$512 to 2048$\times$2048, scale SDXL Turbo~\cite{sauer2023adversarial} from 512$\times$512 to 1024$\times$1024, and scale SDXL~\cite{podell2023sdxl} from 1024$\times$1024 to 4096$\times$4096. Moreover, HiDiffusion is 1.5-2.7$\times$ faster than vanilla Stable Diffusion and is 1.5-6$\times$ faster than previous methods in higher-resolution image generation.
We hope this work can provide valuable guidance for future research on the scalability of diffusion models.

\section{Related Work}

\textbf{High-Resolution Image Synthesis.} 
 The application of diffusion models in high-resolution image generation poses a significant challenge. 
Existing methods have primarily concentrated on diffusion in lower-dimensional spaces (latent diffusion) \cite{rombach2022high}, or divided the generative process into multiple training or finetuning sub-problems \cite{jimenez2023mixture, teng2023relay, hoogeboom2023simple, xie2023difffit}. 
Nevertheless, these solutions render
the diffusion framework is highly intricate. Recently, there has been a growing interest in exploring tuning-free approaches for variable-sized adaptation. \cite{jin2023training} propose an attention scaling factor to improve variable-sized text-to-image synthesis but has not yet addressed the challenge of higher-resolution image generation. MultiDiffusion \cite{bar2023multidiffusion} and SyncDiffusion \cite{lee2023syncdiffusion} manipulated the generation process of a pretrained diffusion model by binding together multiple diffusion generation processes. Despite their advancements, these approaches still exhibit object duplication. Recently, ScaleCrafter~\cite{he2023scalecrafter} mitigates object duplication through re-dilation that can dynamically adjust the convolutional receptive field during inference. ScaleCrafter can effectively address object duplication but somewhat degrades the image quality. DemoFusion~\cite{du2023demofusion} proposed a novel progressive generation schedule with skip residual and dilated sampling. It can generate high-quality high-resolution images, but the long generation time limits its practicality. Different from the previous method, we turn our attention to investigating the properties of the feature map of U-Net. We discover that feature duplication leads to object duplication and propose RAU-Net to effectively resolve it.

\textbf{Diffusion Model Acceleration.} 
As diffusion model training and inference is computationally expensive and time-consuming, particularly in the context of high-resolution images, various methods \cite{chen2023speed, li2023autodiffusion, pan2023effective} have been extensively investigated to accelerate the training and inference of diffusion models.
Unlike fast sampling approaches~\cite{song2020denoising,song2020score,salimans2022progressive, lu2022dpm} that consider using deterministic sampling schemes to improve the sampling speed. ToMeSD ~\cite{bolya2023token} and DeepCache~\cite{ma2023deepcache} speeded up an off-the-shelf diffusion model without training by exploiting natural redundancy in diffusion models. However, they all somewhat compromise image quality. In this paper, through the analysis of the locality of global self-attention in the top blocks, we develop a simple yet effective method MSW-MSA that significantly accelerates the generation of higher-resolution images without the need for fine-tuning and does not compromise image quality.

\section{Method}
\begin{figure}[t]
     \centering
     \begin{subfigure}[b]{0.30\textwidth}
         \centering
         \includegraphics[width=\textwidth]{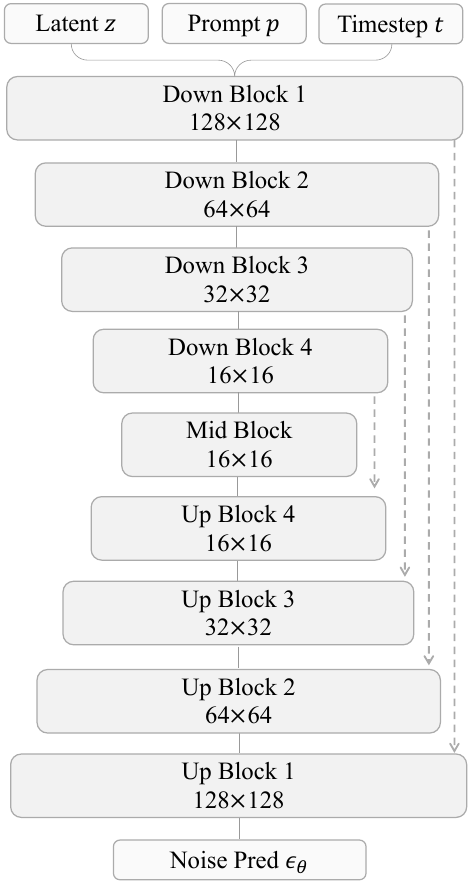}
         \caption{Vanilla U-Net} \label{fig:vanilla_unet}
     \end{subfigure}
     \begin{subfigure}[b]{0.301\textwidth}
         \centering
         \includegraphics[width=\textwidth]{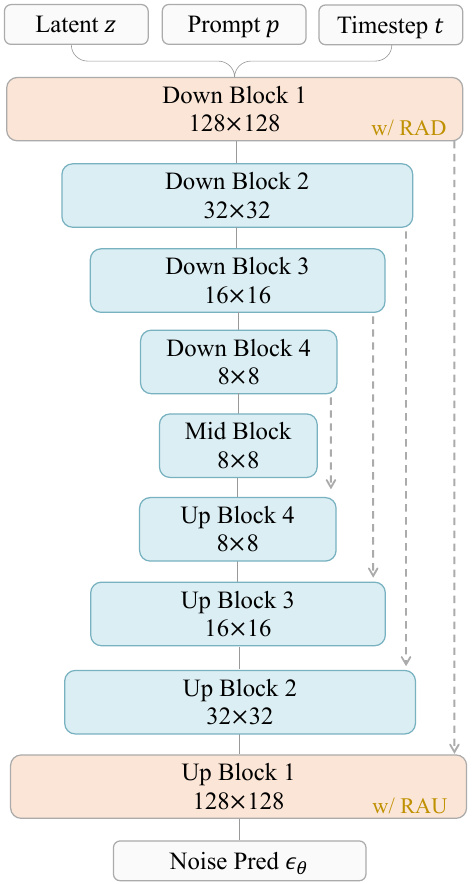}
         \caption{HiDiffusion RAU-Net}
     \end{subfigure}
    \caption{Comparison between vanilla Stable Diffusion’s U-Net architecture and our proposed HiDiffusion RAU-Net architecture on 1024$\times$1024 resolution with SD 1.5~\cite{rombach2022high}. Parameters in all blocks are frozen. The main difference lies in the \textcolor{cyan}{blue} Blocks (differ in the dimensions of feature map) and \textcolor{orange}{orange} Blocks (Our proposed RAD and RAU modules are incorporated into \textcolor{orange}{Block 1}.). } \label{fig:model architecture}
    \label{fig:unet}
    \vspace{-1em}
\end{figure}

\begin{figure}
\centering
\includegraphics[width=0.8\linewidth]{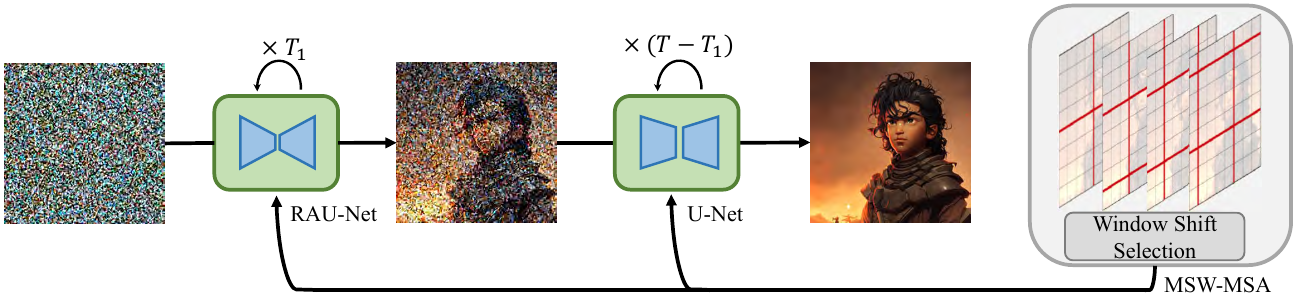}
\caption{The framework of HiDiffusion.}
\label{fig:hidiffusion}
\vspace{-2.5em}
\end{figure}

\subsection{Preliminaries} \label{prelim}
\subsubsection{U-Net architecture.}
\begin{figure}[tb]
  \centering
  \includegraphics[width=\textwidth]{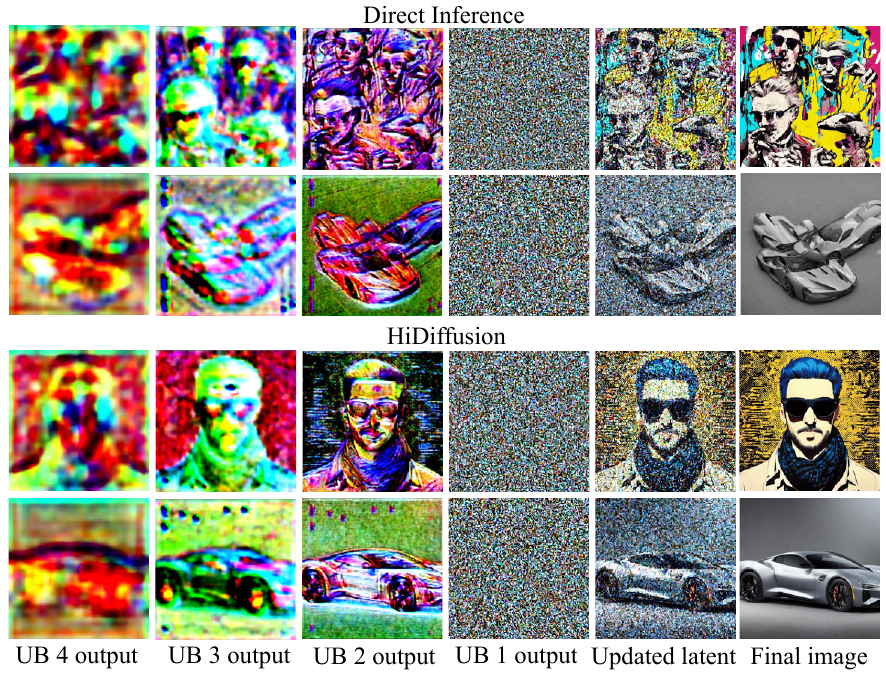}
  \caption{1024$\times$1024 resolution images based on SD 1.5~\cite{rombach2022high}. We visualize the output feature map of U-Net blocks of the 30th step~(50 DDIM steps). UB: Up Block.
  The object structure of the generated images (the last column) is highly correlated with the feature of the deep blocks (UB4, UB3, UB2) in the U-Net. Feature duplication happens when directly generating higher-resolution images and the duplicated features guide the generation direction to object duplication. HiDiffusion can mitigate feature duplication, enabling the generation of reasonable high-resolution images.
  }
  \label{fig:feature_dup}
  \vspace{-1.5em}
\end{figure}

The neural backbone of Stable Diffusion is implemented as a U-Net \cite{dhariwal2021diffusion, ronneberger2015u,rombach2022high}, which contains several Down Blocks, Up Blocks, and a Mid Block, as shown in~\cref{fig:vanilla_unet}. The Mid Block remains unchanged in our method. Consequently, we omit it for the sake of simplicity. Each Down Block and Up Block can be written respectively as:

\vspace{-15pt}
\begin{align}
y & = \mathcal{D}(\mathcal{F}(x,t,p)), \\
y & = \mathcal{U}(\mathcal{F}(x,t,p)),
\vspace{-5pt}
\end{align}
where $x$ is the latent feature, $t$ is the timestep, $p$ is the prompt, 
$\mathcal{F}$ incorporates ResNet~\cite{he2016deep} layers and Vision Transformer~\cite{dosovitskiy2020image} layers, which maintain the dimensions of the feature map. $\mathcal{D}(*)$ represents the downsampler and $\mathcal{U}(*)$ represents the upsampler. $\mathcal{D}(*)$ and $\mathcal{U}(*)$ in vanilla U-Net are computed as:

\vspace{-5pt}
\begin{equation}
    \mathcal{D}(x) = \mathcal{C}_{3,1,2,1}(x),
\end{equation}
\vspace{-10pt}
\begin{equation}
    \mathcal{U}(x) = \mathcal{C}_{3,1,1,1}(interp(x,2)),
\end{equation}
where $\mathcal{C}_{k,p,s,d}$ means convolution filter with kernel size as $k$, padding size as $p$, stride as $s$, dilation rate as $d$. $interp(x,\beta)$ denotes an interpolation function that upsample the resolution by a factor of $\beta$.

\subsubsection{Content generation over timestep.}
The diffusion model progressively performs the denoising process according to the noise schedule. Recent research~\cite{yang2023diffusion,ho2020denoising,choi2022perception,ma2022accelerating,rombach2022high} has found that the diffusion model displays varying denoising behavior over timestep. Diffusion models denoise from structures to details. They generate the low-frequency component in the early denoising stage and the high-frequency component in the late denoising stage.

\subsection{HiDiffusion}

The HiDiffusion framework comprises two components: Resolution-Aware U-Net (RAU-Net) and Modified Shifted Window Multi-head Self-Attention (MSW-MSA). The RAU-Net is designed to overcome object duplication when scaling to higher resolution. MSW-MSA is introduced to improve the inference efficiency of diffusion for higher-resolution image synthesis. The overall framework of HiDiffusion is present in~\cref{fig:hidiffusion}. For each method, we initially present the motivation experiments and then introduce our methods. This section is based on the  1024$\times$1024 resolution image generation with SD 1.5~\cite{rombach2022high}. For other models and extreme resolution, please refer to the appendix for details.

\subsubsection{Resolution-Aware U-Net.} \label{rau-net}

  In this work, we investigate the feature map in the U-Net, aiming to uncover the cause of object duplication and resolve it.

  \noindent\textbf{Obeservation.} \textit{The generated image is highly correlated with the feature map of deep Blocks in structures and feature duplication happens in the deep Blocks.}
  
We provide empirical evidence to demonstrate
this observation in~\cref{fig:feature_dup}.  We discover that the 
the structure of the generated image follows the structural information of the feature maps of the deep blocks. The top block of the U-Net only maps the feature map of deep blocks to noise estimation. We also discover that feature duplication happens in the deep blocks of the U-Net, meaning that the features contain repeated structural information. The highly duplicated features guide the generation direction, resulting in object duplication in the image.

Based on the observation, we aim to reduce the feature duplication in the deep blocks to generate higher-resolution images. As the higher-resolution feature size of deep blocks is larger than the corresponding size in training, these blocks may fail to incorporate feature information globally to generate a reasonable structure. We contend that if the size of the higher-resolution features of deep blocks is reduced to the corresponding size in training, these blocks can generate reasonable structural information and alleviate feature duplication.

Inspired by this motivation, we propose Resolution-aware U-Net~(RAU-Net), a simple yet effective method to dynamically resize the features to match the deep blocks. An illustrative comparison of the vanilla SD 1.5 U-Net and RAU-Net in generating 1024$\times$1024 resolution images is presented in 
\cref{fig:model architecture}. We incorporate our Resolution-Aware Downsapler~(RAD) in Down Block 1 as a substitute for the original downsampler, and likewise, we replace the original upsampler with the Resolution-Aware Upsampler (RAU) in Up Block 1.  RAD downsamples the feature map to guarantee the dimensions of the resulting feature map align with those of the corresponding training images, thereby matching with the deep blocks. On the other hand, RAU simply upsamples the feature size to the desired resolution. Specifically, the RAD and RAU can be written as follows:
\vspace{-4pt}
\begin{equation}
    \mathcal{RAD}(x, \alpha) = \mathcal{R}(\mathcal{C}_{3,1,2,1}(x), \alpha),
\end{equation}
\vspace{-12pt}
\begin{equation}
    \mathcal{RAU}(x, \beta) = \mathcal{C}_{3,1,1,1}(interp(x,\beta)).
\end{equation}

$\alpha$ is the downsampling factor, $\beta$ is the upsampling factor. $\mathcal{R}(*)$ can be achieved by adjusting the conventional downsampler parameters:
\begin{equation}
    \mathcal{R}(\mathcal{C}_{3,1,2,1}(x), \alpha)=\mathcal{C}_{3,p,\alpha,d}(x),
\end{equation}
or achieved by using adaptive pooling:
\begin{equation}
\mathcal{R}(\mathcal{C}_{3,1,2,1}(x), \alpha)=ada\_pool(\mathcal{C}_{3,1,2,1}(x), \frac{\alpha}{2}),
\end{equation}
We mainly choose the first variant in this paper. We conduct ablations about RAD in the appendix. 
RAU can be simply achieved with an upscale interpolation such as bilinear interpolation.
For 1024$\times$1024 image generation, we need to downsample the feature map by a factor of 4 to match the deep blocks, i.e., $\alpha=4$. This downsampling factor is twice the downsampling factor of the conventional downsampler. We set $d=2$, $p=2$. For RAU, we only need to adjust the interpolation factor to 4. Compared to the original samplers, both RAD and RAU do not introduce additional trainable parameters. Therefore, RAD and RAU can be integrated into vanilla U-Net without further fine-tuning.

Upon incorporating RAU-Net into SD 1.5~\cite{rombach2022high}, we address the object duplication problem. But we also bring blurry images with bad details~(please refer to the appendix). As mentioned in~\ref{prelim}, diffusion models denoise from structures to details.
Our RAU-Net introduces additional downsampling and upsampling operations, leading to a certain degree of information loss. In the early stages of denoising, RAU-Net can generate reasonable structures with minimal impact from information loss. However, in the later stages of denoising when generating fine details, the information loss in RAU-Net results in the loss of image details and a degradation in quality. 
 Consequently, we establish a \textbf{Switching Threshold} $T_1$, such that when the denoising steps $t < T_1$, RAU-Net is employed, conversely, when the denoising steps $t \geq T_1$, vanilla U-Net is utilized. This simple adjustment can effectively counteract the information loss brought about by RAU-Net, significantly improving the image quality. Moreover, we observed that the parameter $T_1$ is not sensitive, with settings between 20 and 40 for 1024 $\times$ 1024 generation yielding notably superior performance with 50 DDIM steps. Please refer to the appendix for details.

\subsubsection{Modified Shifted Window Attention}

\begin{figure}[t]
	\centering 
	
	\begin{subfigure}[t]{0.3\textwidth}
		\centering
		\includegraphics[width=\textwidth]{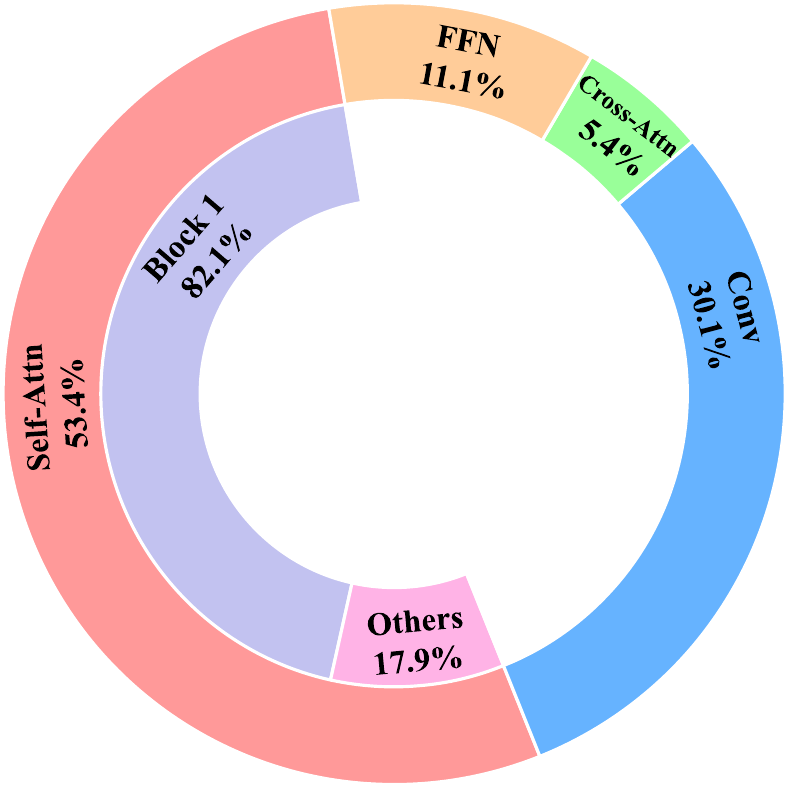}
		\caption{Operation consumption.}
		\label{fig:time_proportion}
	\end{subfigure}
	\begin{subfigure}[t]{0.58\textwidth}
		\centering
		\includegraphics[width=\textwidth]{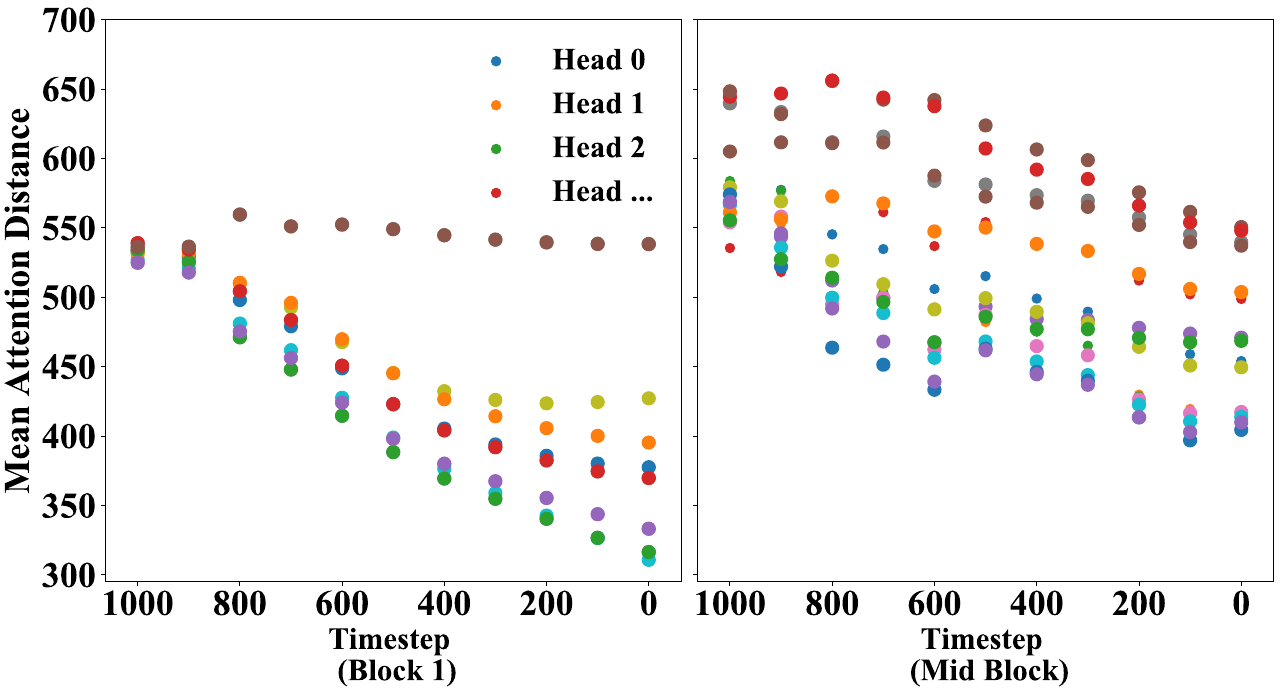}
		\caption{Mean attention distance.}
		\label{fig:mean_dis}
	\end{subfigure}
\vspace{-0.3cm}
	\caption{Analysis of the time consumption and mean attention distance. (a) The self-attention operation of Block 1 significantly dominates the time consumption. (b) A pronounced locality is evident in the self-attention mechanism of the top blocks.} 
	\label{fig:focal_ohem} 
\vspace{-2em}
\end{figure} 

Stable Diffusion combined with RAU-Net is capable of generating higher-resolution images with high quality. However, it still faces an efficiency challenge: unaffordable slow speed in generating high-resolution images. In this section, we revisit the consumption and properties of operations in U-Net, trying to accelerate the diffusion model.

  \noindent\textbf{Obeservation.} \textit{The self-attention of the top blocks takes the dominant consumption. However, it demonstrates locality.}

Given a latent feature map with 128$\times$128 (corresponding to 1024$\times$1024 resolution in pixel space), ~\cref{fig:time_proportion} shows the time consumption of each operation in SD 1.5. It can be observed that self-attention, especially in Block 1~(i.e. Down Block 1 and Up Block 1), takes the dominant consumption. Driven by previous local self-attention works in vision~\cite{liu2021swin,pan2023slide}, we visualize the mean attention 
distance for each head across different timesteps of Block 1~(top block) and Mid Block~(deep block), as shown in~\cref{fig:mean_dis}. We surprisingly find that the self-attention mechanism in the top blocks demonstrates a pronounced locality. Certain heads are observed to attend to approximately half of the image, while others focus on even more confined regions close to the query location. 

According to this observation, it is suggested to propose local self-attention for efficient computation. We delve into how to design local attention to ensure acceleration while maintaining image quality. Based on window attention~\cite{liu2021swin}, we propose Modified Shifted Window Attention~(MSW-MSA), a simple yet effective approach for lossless acceleration to replace the original global attention. Specifically, MSW-MSA has two modifications: \textbf{(\romannumeral1) Large window attention}.~\cref{fig:mean_dis} indicates that the mean attention distance of the top blocks is local but not very small. This suggests that the small window attention utilized in vision field~\cite{liu2021swin} may not be suitable for the diffusion model. To achieve a balance between acceleration and image quality, we opt for a larger window attention. The experimentally validated window size is $(H/2, W/2)$, where $H$ and $W$ respectively represent the height and width of the input feature. \textbf{(\romannumeral2) Shift window operation across timestep}. Window shift operation is needed to introduce cross-window connections. However, Stable Diffusion transformer block has only one self-attention module that is unable to process window attention and shift window attention successively. To apply shift window operation, we propose to shift different strides based on the timesteps. Specifically, we adopt a random selection strategy, where at each timestamp, we randomly select a stride parameter from a fixed set of shift strides. This approach enables the integration of information from diverse windows. Our MSW-MSA can be written as:
\begin{equation}
y = \text{MSW-MSA}(x,w,s(t))+x,
\label{eq:wmsa}
\end{equation}
where $w$ is the window size, $s(t)$ is the shifted stride function dependant on the timestep $t$.

We substitute the global self-attention in Block 1 with MSW-MSA. It is worth noting that while other blocks can integrate MSW-MSA, the resulting efficiency gains are not substantial. Experiments demonstrate that our MSW-MSA approach can significantly reduce time consumption \textbf{without compromising image quality} in higher-resolution image synthesis.

\begin{table}
\begin{center}
\begin{footnotesize}
 \begin{adjustbox}{max width=\linewidth}
  \begin{tabular}[h]{lcccccccc}
   \toprule
    & &  &\multicolumn{3}{c}{ImageNet}& \multicolumn{3}{c}{COCO} \\
    \cmidrule(lr){4-6} \cmidrule(lr){7-9}
    Method & Resolution & Latency (s) $\downarrow$ & FID $\downarrow$ & pFID $\downarrow$ & CLIP $\uparrow$ & FID $\downarrow$ & pFID $\downarrow$ & CLIP $\uparrow$\\
    \midrule
    SD 1.5 & \multirow{6}{*}{1024 $\times$ 1024}& 16.23  &25.55 & 36.36 & 0.295& 38.21 & 49.20 &0.309\\
    SD 1.5 + HiDiffusion & & \textbf{8.26(1.96$\times$)} & \textbf{21.81} & \textbf{30.86} &\textbf{0.307} & \textbf{21.36} & \textbf{31.59} &\textbf{0.323}\\
    \cmidrule(lr){1-1}
    SD 2.1 & & 12.99  & 24.63  & 36.15 & 0.299& 31.33 & 37.43 & 0.314\\
    SD 2.1 + HiDiffusion & & \textbf{7.33(1.77$\times$)} & \textbf{22.34} & \textbf{32.71} & \textbf{0.309} & \textbf{20.77}& \textbf{31.51} & \textbf{0.326} \\
    \cmidrule(lr){1-1}
    SDXL Turbo & & 5.72 & 74.23  & 76.08 &  0.300& 23.45 & 35.10 & 0.325\\
    SDXL Turbo+ HiDiffusion & & \textbf{4.65(1.23$\times$)} & \textbf{27.76} & \textbf{32.63} & \textbf{0.317}  & \textbf{20.89}  &  \textbf{32.90}& \textbf{0.330}  \\

    \midrule
    SD 1.5 & \multirow{6}{*}{2048 $\times$ 2048} & 165.76& 53.03 &  35.96 & 0.284  &  78.53 & 42.82 & 0.286  \\
    SD 1.5 + HiDiffusion & & \textbf{58.38(2.83$\times$)} & \textbf{27.33} & \textbf{33.42} & \textbf{0.307} & \textbf{28.93} & \textbf{34.70}  &\textbf{0.321} \\
 \cmidrule(lr){1-1}
    SD 2.1 & & 118.32 & 60.60& 41.64 & 0.281 & 82.74 &  47.62 &0.289\\
    SD 2.1 + HiDiffusion & & \textbf{45.33(2.61$\times$)} & \textbf{30.67} &  \textbf{37.14} & \textbf{0.305}& \textbf{32.87}& \textbf{35.76} & \textbf{0.320}\\
    \cmidrule(lr){1-1}
    SDXL & & 84.24 & 27.48 &  30.67& 0.300 & 28.71 & 32.44 & 0.318 \\
    SDXL + HiDiffusion & & \textbf{53.29(1.58$\times$)} & \textbf{22.22} & \textbf{28.27} & \textbf{0.314} & \textbf{20.89} & \textbf{29.21} &\textbf{0.332}  \\
    \midrule
    SDXL$^{\dag}$ & \multirow{2}{*}{4096 $\times$ 4096} & 769.65 & 118.30 & 89.96 & 0.277  & - & - & - \\
    SDXL + HiDiffusion$^{\dag}$ & & \textbf{286.97(2.68$\times$)} & \textbf{64.12} & \textbf{74.91} & \textbf{0.299}  & - & - &- \\    
    \bottomrule
\end{tabular}
 \end{adjustbox}
\end{footnotesize}
 \caption{Comparison of vanilla Stable Diffusion and our Hidiffusion in zero-shot text-guided image synthesis on ImageNet and COCO dataset. $^{\dag}$ means we generate 1K images for quantitative evaluation due to the heavy computational burden.}
 \label{tab:fid_clip}
\end{center}
\vspace{-4em}
\end{table}

\section{Experiments}

\subsection{Experiment Settings}
\label{sec:exp_settings}

\begin{figure}[tb]
  \centering
  \includegraphics[width=\textwidth]{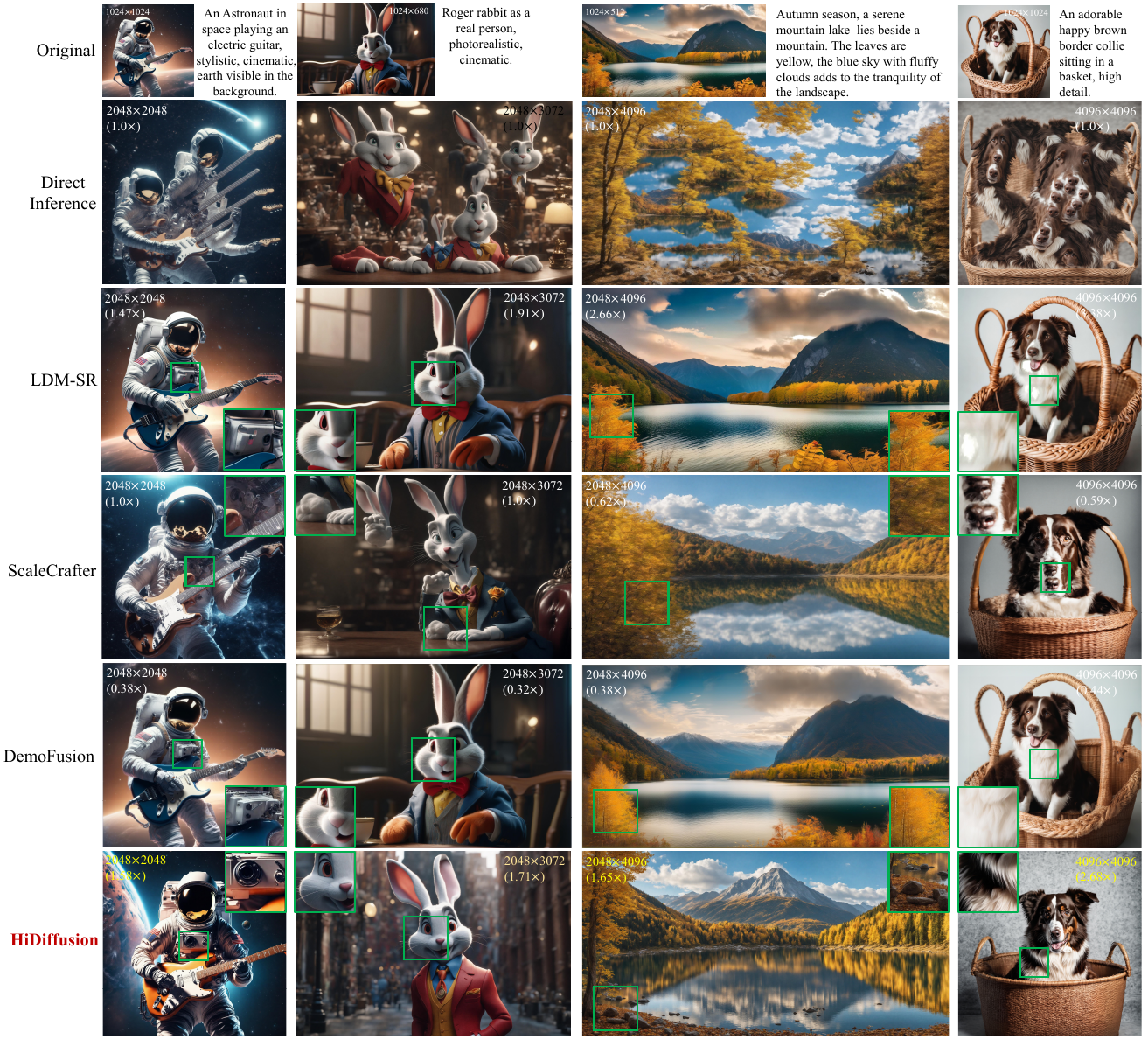}
  \caption{Qualitative comparison with other methods based on SDXL~\cite{podell2023sdxl}. The input prompt is located to the right of the original image. The first line of text in the image indicates the image resolution, while the second line indicates the inference speed relative to direct inference. Best viewed when zoomed in.
  }
  \label{fig:sample_exp}
\vspace{-2em}
\end{figure}

In this work, we evaluate the performance of our HiDiffusion on SD 1.5~\cite{rombach2022high}, SD 2.1~\cite{rombach2022high}, SDXL Turbo~\cite{sauer2023adversarial} and SDXL~\cite{podell2023sdxl}. We apply our approach to text-guided image synthesis on higher resolution ranging from 4$\times$ to even 16$\times$ times the training image resolution. For quantitative evaluation, we use Frechet Inception Distance~(FID)~\cite{heusel2017gans} to measure the realism of the output distribution. FID downsamples all images to a common size of 299, which ignores high-resolution
details. We further use patches FID~(pFID)~\cite{chai2022any} to evaluate image details. CLIP Score~\cite{radford2021learning} is proposed to evaluate the alignment between image and text. We compare our HiDiffusion with other methods on ImageNet~\cite{russakovsky2015imagenet} and COCO~\cite{lin2014microsoft} datasets. Without further elaboration, we generate 10K~(10 per class)  images to compute metrics for ImageNet evaluation and generate 40,504 (1 caption per image) images from COCO 2014 validation captions to compute metrics for COCO evaluation. We use xFormers~\cite{xFormers2022} by default. The model latency is measured on a single NVIDIA V100 with a batch size of 1.

We introduce the parameter setting of SD 1.5, please refer to the appendix for the parameter setting of other models. For 1024$\times$1024 generation, we incorporate RAD and RAU in Block 1 and set $\alpha=\beta=4$. We set the window size as $(64, 64)$. The predefined set of shift strides is $\{(0,0), (16,16),(32,32),(48,48)\}$. All experiments are conducted with 50 DDIM steps. The classifier-free guidance scale is 7.5. The switching threshold $T_1$ is set as $20$. When extended to 2048$\times$2048, we can simply set $\alpha=8$. However, a sharp change in resolution caused by interpolation may bring blurriness, hence we adopt a progressive approach by incorporating RAU and RAD with $\alpha=\beta=4$ into Block 1 and Block 2, respectively, This allows the feature map to gradually match the deep blocks. Please refer to the appendix for more details.

\begin{table}
\begin{center}
\begin{footnotesize}
 \begin{adjustbox}{max width=\linewidth}
  \begin{tabular}[h]{lccccccccc}
   \toprule 
    & & &  &\multicolumn{3}{c}{ImageNet}& \multicolumn{3}{c}{COCO} \\
    \cmidrule(lr){5-7} \cmidrule(lr){8-10}
    Backbone & Method & Resolution & Latency (s) $\downarrow$ & FID $\downarrow$ & pFID $\downarrow$ & CLIP $\uparrow$ & FID $\downarrow$ & pFID $\downarrow$ & CLIP $\uparrow$\\
    \midrule
    \multirow{2}{*}{SD 1.5} & ScaleCrafter~\cite{he2023scalecrafter} & \multirow{4}{*}{1024 $\times$ 1024} & 17.94  & 54.90 & 74.39 & 0.302 & 81.68 &  87.32& 0.318 \\
    &HiDiffusion~(ours) &  &  \textbf{8.26} &  \textbf{50.14} & \textbf{65.80} & \textbf{0.307} & \textbf{80.73} & \textbf{84.19}& \textbf{0.322}\\
    \cmidrule(lr){1-2}
    \multirow{2}{*}{SD 2.1}&ScaleCrafter~\cite{he2023scalecrafter} & & 14.58 & 57.87  & 75.39 & 0.303 & 80.02 &84.69 & 0.321\\
    &HiDiffusion~(ours) & & \textbf{7.33} & \textbf{50.20} & \textbf{68.69} & \textbf{0.309} & \textbf{79.19} & \textbf{82.47} & \textbf{0.325}\\

    \midrule

    \multirow{2}{*}{SD 1.5} & ScaleCrafter~\cite{he2023scalecrafter} & \multirow{7}{*}{2048 $\times$ 2048} &   287.89 & 65.04  & 86.83 & 0.299 & 98.30 & 104.49 & 0.309 \\
    &HiDiffusion~(ours) & & \textbf{58.38} &  \textbf{53.35} &  \textbf{65.95} & \textbf{0.307} & \textbf{86.15} & \textbf{86.82} & \textbf{0.319} \\
    \cmidrule(lr){1-2}
    \multirow{2}{*}{SD 2.1}&ScaleCrafter~\cite{he2023scalecrafter} & & 216.48 & 78.12  & 102.14 & 0.295 & 110.75 & 122.49& 0.299 \\
    &HiDiffusion~(ours) & & \textbf{45.33} & \textbf{57.71}& \textbf{72.09} & \textbf{0.306} & \textbf{88.56} &  \textbf{87.07} & \textbf{0.317} \\
    \cmidrule(lr){1-2}
    \multirow{3}{*}{SDXL}&ScaleCrafter~\cite{he2023scalecrafter} & & 85.83 & 49.97  & 72.64 & 0.310 & \textbf{82.14} &  90.45 & 0.329 \\
    &DemoFusion~\cite{du2023demofusion} & & 222.79& 48.36 &  66.41 & 0.311 & 85.92  & 85.59 & 0.331 \\
    &HiDiffusion~(ours) & & \textbf{53.29} & \textbf{47.01} & \textbf{62.29} & \textbf{0.315} & 84.66 & \textbf{84.58} & \textbf{0.333}\\
    
    \midrule

    \multirow{3}{*}{SDXL} & ScaleCrafter~\cite{he2023scalecrafter} & \multirow{3}{*}{4096 $\times$ 4096} & 1298.39& 78.90 & 102.63  & 0.305 & - &  -& - \\
    &DemoFusion~\cite{du2023demofusion} & & 1735.58 & \textbf{58.93} & 76.53 & \textbf{0.311}& - & - & - \\
    &HiDiffusion~(ours) & & \textbf{286.97} & 64.12 & \textbf{74.91} & 0.307 & - & - & -\\
    \bottomrule
\end{tabular}
 \end{adjustbox}
\end{footnotesize}
 \caption{Comparison of high-resolution generation method and our HiDiffusion in zero-shot text-guided image synthesis on ImageNet and COCO dataset.}
 \label{tab:fid_clip_md}
\end{center}\vspace{-2em}
\end{table}

\subsection{Main results}
\label{sec:text_guided}

In this section, We incorporate our method into SD 1.5~\cite{rombach2022high}, SD 2.1~\cite{rombach2022high}, SDXL Turbo~\cite{sauer2023adversarial} and SDXL~\cite{podell2023sdxl} to evaluate the effectiveness of our method. SD 1.5 and SD 2.1 are capable of generating images with 512$\times$512 resolution. We integrate HiDiffusion into them to scale the resolution to 1024$\times$1024 and 2048$\times$2048. We use HiDiffusion to scale the generation resolution of SDXL Turbo to 1024$\times$1024. For SDXL, which is trained for generating 1024$\times$1024 images, we incorporate our method to scale the resolution to 2048$\times$2048 and 4096$\times$4096. Besides fixed aspect ratios, we also generate images with various aspect ratios, such as 512$\times$2048, 1280$\times$1024 and 2048$\times$4096, and so on, please refer to the appendix. We compare our method with the high-resolution generation method ScaleCrafter~\cite{he2023scalecrafter} and DemoFusion~\cite{du2023demofusion}. For the acceleration of diffusion model, we compare the diffusion acceleration method Token Merge for Stable Diffusion (ToMeSD)~\cite{bolya2023token} and DeepCache~\cite{ma2023deepcache} with our proposed MSW-MSA. Moreover, we compare our method with super-resolution method for a thorough evaluation, even though the latter requires a large number of high-resolution images and extra training efforts to train a super-resolution model.

\textbf{Comparision with vanilla SD.} In~\cref{fig:sample_exp}, we show  qualitative comparison between the vanilla SD~(direct inference) and our method. It can be easily seen the vanilla SD suffers from object duplication and degradation in visual quality as well. In contrast, our HiDiffusion mitigates the duplication problem and holds more realistic image structures simultaneously. 
The quantitative results are shown in~\cref{tab:fid_clip}. Our approach outperforms vanilla SD in both quality and image-text alignment. We achieve much better metric scores across all experiment settings, especially for the images with much higher resolution (a significant FID improvement from 78.53 to 28.93 for SD 1.5 on COCO dataset with 2048$\times$2048 resolution). Furthermore, HiDiffusion significantly accelerates diffusion inference. For instance, when incoporating HiDiffusion, SDXL is 2.68$\times$ faster than the vanilla model when generating images with 4096$\times$4096 resolution.

\begin{table}\vspace{1em}
\begin{center}
\begin{footnotesize}
 \begin{adjustbox}{max width=\linewidth}
  \begin{tabular}[h]{lccccc}
   \toprule

    Method & Resolution & Latency (s) $\downarrow$ & FID $\downarrow$ & pFID $\downarrow$ & CLIP $\uparrow$\\
    \midrule
    SD 1.5 + LDM-SR$^*$ &  \multirow{4}{*}{1024 $\times$ 1024} & 18.61 & \textbf{17.56}  &  36.39& \textbf{0.307} \\
    SD 1.5 + HiDiffusion & & \textbf{8.26(2.25$\times$)} & 21.81  & \textbf{30.86}& \textbf{0.307} \\
    SD 2.1 + LDM-SR$^*$ & & 18.48 & \textbf{18.54} & 37.99 & 0.308\\
    SD 2.1 + HiDiffusion & & \textbf{7.33(2.52$\times$)} & 22.34  & \textbf{32.71} & \textbf{0.309} \\

    \midrule
    SD 1.5 + LDM-SR$^*$ &  \multirow{6}{*}{2048 $\times$ 2048} & 73.03 & \textbf{17.32} & 35.57 & \textbf{0.308} \\
    SD 1.5 + HiDiffusion & & \textbf{58.28(1.25$\times$)} & 27.33  & \textbf{33.42} & 0.307   \\
    SD 2.1 + LDM-SR$^*$ & &72.90 & \textbf{18.16}  & 41.87 & \textbf{0.308} \\
    SD 2.1 + HiDiffusion & & \textbf{45.33(1.61$\times$)} & 30.67  & \textbf{37.14}  & 0.304 \\ 

    SDXL + LDM-SR$^*$ & & 57.28 &  \textbf{46.73} & 68.98 & \textbf{0.315} \\
    SDXL + HiDiffusion & & \textbf{53.29(1.07$\times$)}  & 47.01  & \textbf{62.29}  & \textbf{0.315} \\ 
    \midrule
    SDXL + LDM-SR$^*$ & \multirow{2}{*}{4096 $\times$ 4096} & \textbf{227.57}  & \textbf{57.04}  &  79.66&  \textbf{0.315}\\
    SDXL + HiDiffusion & & 286.97  & 64.12  & \textbf{74.91}  & 0.307 \\ 
    \bottomrule
\end{tabular}
 \end{adjustbox}
\end{footnotesize}
 \caption{Comparison of diffusion super-resolution and our method in zero-shot text-guided image synthesis on ImageNet dataset. $^*$ is a two-stage method, requiring extra high-resolution datasets and training efforts to train a large super-resolution model. Our approach is one-stage and can generate high-resolution images without any extra high-resolution data collection and training costs. }
 \label{tab:fid_clip_sr}
\end{center}\vspace{-2.5em}
\end{table}

\textbf{Comparison with high-resolution synthesis method.} We present a qualitative comparison between ScaleCrafter~\cite{he2023scalecrafter}, DemoFusion~\cite{du2023demofusion} and our method in~\cref{fig:sample_exp}. We observe that all three methods can generate reasonable structures. But our method can generate much richer details than ScaleCrafter and DemoFusion.
~\cref{tab:fid_clip_md} shows quantitative results. Note that we generate 1K images for both ImageNet and COCO quantitative evaluation in this section
due to the heavy computational burden. our method outperforms ScaleCrafter across almost all models and achieves comparable or better performance than DemoFusion. It is worth noting that we significantly surpass ScaleCrafter and DemoFusion in generation efficiency: HiDiifusion is 2-5$\times$ faster than ScaleCrafter and is 4-6$\times$ faster than DemoFusion.

\textbf{Comparison with diffusion super-resolution models.}
Instead of directly generating high-resolution images using a single diffusion model, a more commonly used approach in the community is to generate images with original resolution using Stable Diffusion and scale them to higher resolution using an extra super-resolution model. Although this approach requires additional high-resolution training datasets and extensive training efforts to train a large super-resolution model, we compare it for a thorough comparison, despite the inherent unfairness to our one-stage and training-free method. We compare our method with a pretrained Stable Diffusion super-resolution
model LDM-SR~\cite{rombach2022high}.
~\cref{tab:fid_clip_sr} shows the quantitative results. In terms of generation efficiency, our method outperforms LDM-SR significantly when based on SD 1.5 and SD 2.1. On SDXL, our speed is roughly on par with that of LDM-SR. Both LDM-SR and our method are capable of generating plausible structures. However, our method exhibits lower pFID. This indicates that the images generated by our method are more detailed. We visualize the synthesized samples in~\cref{fig:sample_exp}. Compared with LDM-SR, a distinction can be observed in terms of visual image detail quality. Our method directly generates content on a 2048$\times$2048 or 4096$\times$4096 canvas, resulting in higher richness, sharper characteristics, and fine-grained details.

\begin{table}[]
    \centering
    \resizebox{0.9\linewidth}{!}{%
    \begin{tabular}{l c c c c c}
        \toprule
         Method      & Resolution                 & Latency (s) & FID $\downarrow$ & pFID $\downarrow$ & CLIP $\uparrow$ \\
         \midrule
        Baseline &      \multirow{5}{*}{1024$\times$1024}                      &          14.31   &  22.93   &  32.80& \textbf{0.307}   \\
        ToMeSD~\cite{bolya2023token} &                          &    8.73          &   22.76  & 35.23  & 0.305         \\
        DeepCache~(interval=3)~\cite{ma2023deepcache} &                          &   \textbf{6.62}           & 25.31  &  35.61  & 0.306        \\
        DeepCachee~(interval=2)~\cite{ma2023deepcache} &                          &    8.49          & 24.07  &   33.89 & 0.306       \\
        MSW-MSA~(ours)   &                            &       8.26     &   \textbf{21.80}  & \textbf{30.86}  &  \textbf{0.307}        \\
        \midrule

        Baseline &      \multirow{5}{*}{2048$\times$2048}                      &        151.58     &   28.21  & 33.79 & 0.306   \\
        ToMeSD~\cite{bolya2023token} &                          &     67.02         &   27.49 & 34.38&   0.305        \\
        DeepCache~(interval=3)~\cite{ma2023deepcache} &                          &   70.70        &  34.49  & 42.13  & 0.305     \\
        DeepCache~(interval=2)~\cite{ma2023deepcache} &                          &    90.20       &   32.72 &  41.44 &    \textbf{0.307}      \\
        MSW-MSA~(ours)   &                            &        \textbf{58.38}    &  \textbf{27.33}   & \textbf{33.42} &  \textbf{0.307}        \\

        \bottomrule
    \end{tabular}
    }
     \caption{Quantitative evaluation of diffusion acceleration methods and our proposed MSW-MSA in zero-shot text-guided image synthesis on ImageNet. Baseline indicates SD 1.5 with RAU-Net.
        }
    \label{table:tome_window}
\end{table}

\textbf{Comparison with diffusion acceleration method.}
We compare our method with ToMeSD~\cite{bolya2023token} and DeepCache~\cite{ma2023deepcache}. The comparison is implemented at the resolutions of 1024$\times$1024 and 2048$\times$2048 based on SD 1.5 with RAU-Net. ~\cref{table:tome_window} shows the quantitative results on ImageNet~\cite{russakovsky2015imagenet}. As observed, our proposed MSW-MSA outperforms ToMeSD and DeepCache across almost all metrics. Although ToMeSD and DeepCache have achieved significant acceleration, they have also compromised image quality. In contrast, our MSW-MSA can achieve or even surpass the acceleration effects of ToMeSD and DeepCache 
without compromising image quality. Please refer to the appendix for the visual sample comparison.

\begin{table}[]
    \centering
    \resizebox{0.6\linewidth}{!}{%
    \begin{tabular}{c c c c c c}
        \toprule
         RAU-Net      & MSW-MSA                 & Latency (s) & FID $\downarrow$ & pFID $\downarrow$ & CLIP $\uparrow$ \\
         \midrule
         &                        &     16.23      &   25.55 & 36.36 &  0.295  \\
        \checkmark &                        &         14.31  &  22.93  & 32.80  & 0.307   \\
         &  \checkmark                       &         10.15  &  23.28 &  33.84 &  0.297  \\
         \checkmark  &  \checkmark                       &        \textbf{8.26}  &  \textbf{21.81}   & \textbf{30.86}  &  \textbf{0.307}  \\
        \bottomrule
    \end{tabular}
    }
     \caption{Ablation about the components of HiDiffusion with SD 1.5 in zero-shot text-guided image synthesis on ImageNet.
        \vspace{-1em}}
    \label{table:hi_ablation}
\end{table}

\subsection{Ablation study}

We ablate the components of HiDiffusion on 1024$\times$1024 resolution image generation based on SD 1.5.~\cref{table:hi_ablation} shows quantitative results of all possible combinations. It can be seen that both RAU-Net and MSW-MSA bring improvements in performance and speed. 
RAU-Net uses RAD to adjust the feature size to the training dimensions, resulting in speed benefits. Meanwhile, the window attention in MSW-MSA prevents an overwhelming number of tokens from global interaction, thereby reducing token homogenization and enhancing performance. This perspective is discussed in~\cite{jin2023training}.
When both are used simultaneously, the optimal result can be achieved. We provide visual samples in the appendix.

\section{Conclusion}
In this paper, we propose a tuning-free framework named HiDiffusion for higher-resolution image generation. HiDiffusion includes Resolution-Aware U-Net~(RAU-Net) that makes higher-resolution generation possible and Modified Shifted Window Multi-head Self-Attention~(MSW-MSA) that makes higher-resolution generation efficient. Empirically,  HiDiffusion can be incorporated into SD 1.5~\cite{rombach2022high}, 2.1~\cite{rombach2022high}, XL~\cite{podell2023sdxl}, and XL Turbo~\cite{sauer2023adversarial}, and scale them to generate 1024$\times$1024, 2048$\times$2048, or even 4096$\times$4096 resolution images, while significantly reducing inference time. Compared to previous higher-resolution image generation methods, we can generate images with richer details in less inference time. 
We hope our work can bring insight to future works about the scalability of diffusion models.

\textbf{Limitations and future work:} Our approach involves directly harnessing the intrinsic potential of Stable Diffusion without any additional training or fine-tuning, hence some inherent issues posed by stable diffusion persist, such as the requirement for prompt engineering to obtain more promising images. Furthermore, we can explore better ways to integrate with super-resolution models to achieve higher resolution and amazing image generation outcomes.

%
%
\bibliographystyle{splncs04}
\bibliography{main}

\begin{thebibliography}{10}
\providecommand{\url}[1]{\texttt{#1}}
\providecommand{\urlprefix}{URL }
\providecommand{\doi}[1]{https://doi.org/#1}

\bibitem{bar2023multidiffusion}
Bar-Tal, O., Yariv, L., Lipman, Y., Dekel, T.: Multidiffusion: Fusing diffusion paths for controlled image generation. In: ICML (2023)

\bibitem{bolya2023token}
Bolya, D., Hoffman, J.: Token merging for fast stable diffusion. In: CVPRW. pp. 4598--4602 (2023)

\bibitem{chai2022any}
Chai, L., Gharbi, M., Shechtman, E., Isola, P., Zhang, R.: Any-resolution training for high-resolution image synthesis. In: ECCV. pp. 170--188. Springer (2022)

\bibitem{chen2023speed}
Chen, Y.H., Sarokin, R., Lee, J., Tang, J., Chang, C.L., Kulik, A., Grundmann, M.: Speed is all you need: On-device acceleration of large diffusion models via gpu-aware optimizations. In: CVPR. pp. 4650--4654 (2023)

\bibitem{choi2022perception}
Choi, J., Lee, J., Shin, C., Kim, S., Kim, H., Yoon, S.: Perception prioritized training of diffusion models. In: CVPR. pp. 11472--11481 (2022)

\bibitem{dhariwal2021diffusion}
Dhariwal, P., Nichol, A.: Diffusion models beat gans on image synthesis. vol.~34, pp. 8780--8794 (2021)

\bibitem{dosovitskiy2020image}
Dosovitskiy, A., Beyer, L., Kolesnikov, A., Weissenborn, D., Zhai, X., Unterthiner, T., Dehghani, M., Minderer, M., Heigold, G., Gelly, S., et~al.: An image is worth 16x16 words: Transformers for image recognition at scale. In: ICLR (2021)

\bibitem{du2023demofusion}
Du, R., Chang, D., Hospedales, T., Song, Y.Z., Ma, Z.: Demofusion: Democratising high-resolution image generation with no \$\$\$. In: CVPR (2024)

\bibitem{he2016deep}
He, K., Zhang, X., Ren, S., Sun, J.: Deep residual learning for image recognition. In: CVPR. pp. 770--778 (2016)

\bibitem{he2023scalecrafter}
He, Y., Yang, S., Chen, H., Cun, X., Xia, M., Zhang, Y., Wang, X., He, R., Chen, Q., Shan, Y.: Scalecrafter: Tuning-free higher-resolution visual generation with diffusion models. In: ICLR (2024)

\bibitem{heusel2017gans}
Heusel, M., Ramsauer, H., Unterthiner, T., Nessler, B., Hochreiter, S.: Gans trained by a two time-scale update rule converge to a local nash equilibrium. In: NeurIPS (2017)

\bibitem{ho2020denoising}
Ho, J., Jain, A., Abbeel, P.: Denoising diffusion probabilistic models. In: NeurIPS. pp. 6840--6851 (2020)

\bibitem{hoogeboom2023simple}
Hoogeboom, E., Heek, J., Salimans, T.: simple diffusion: End-to-end diffusion for high resolution images. arXiv preprint arXiv:2301.11093  (2023)

\bibitem{jimenez2023mixture}
Jim{\'e}nez, {\'A}.B.: Mixture of diffusers for scene composition and high resolution image generation. arXiv preprint arXiv:2302.02412  (2023)

\bibitem{jin2023training}
Jin, Z., Shen, X., Li, B., Xue, X.: Training-free diffusion model adaptation for variable-sized text-to-image synthesis. arXiv preprint arXiv:2306.08645  (2023)

\bibitem{lee2023syncdiffusion}
Lee, Y., Kim, K., Kim, H., Sung, M.: Syncdiffusion: Coherent montage via synchronized joint diffusions. In: NeurIPS (2023)

\bibitem{xFormers2022}
Lefaudeux, B., Massa, F., Liskovich, D., Xiong, W., Caggiano, V., Naren, S., Xu, M., Hu, J., Tintore, M., Zhang, S., Labatut, P., Haziza, D.: xformers: A modular and hackable transformer modelling library. \url{https://github.com/facebookresearch/xformers} (2022)

\bibitem{li2023autodiffusion}
Li, L., Li, H., Zheng, X., Wu, J., Xiao, X., Wang, R., Zheng, M., Pan, X., Chao, F., Ji, R.: Autodiffusion: Training-free optimization of time steps and architectures for automated diffusion model acceleration. pp. 7105--7114 (2023)

\bibitem{lin2014microsoft}
Lin, T.Y., Maire, M., Belongie, S., Hays, J., Perona, P., Ramanan, D., Doll{\'a}r, P., Zitnick, C.L.: Microsoft coco: Common objects in context. In: ECCV. pp. 740--755 (2014)

\bibitem{liu2021swin}
Liu, Z., Lin, Y., Cao, Y., Hu, H., Wei, Y., Zhang, Z., Lin, S., Guo, B.: Swin transformer: Hierarchical vision transformer using shifted windows. In: ICCV. pp. 10012--10022 (2021)

\bibitem{lu2022dpm}
Lu, C., Zhou, Y., Bao, F., Chen, J., Li, C., Zhu, J.: Dpm-solver: A fast ode solver for diffusion probabilistic model sampling in around 10 steps. In: NeurIPS. pp. 5775--5787 (2022)

\bibitem{lu2022dpm++}
Lu, C., Zhou, Y., Bao, F., Chen, J., Li, C., Zhu, J.: Dpm-solver++: Fast solver for guided sampling of diffusion probabilistic models. arXiv preprint arXiv:2211.01095  (2022)

\bibitem{ma2022accelerating}
Ma, H., Zhang, L., Zhu, X., Feng, J.: Accelerating score-based generative models with preconditioned diffusion sampling. In: ECCV. Springer (2022)

\bibitem{ma2023deepcache}
Ma, X., Fang, G., Wang, X.: Deepcache: Accelerating diffusion models for free. In: CVPR (2024)

\bibitem{meng2023distillation}
Meng, C., Rombach, R., Gao, R., Kingma, D., Ermon, S., Ho, J., Salimans, T.: On distillation of guided diffusion models. In: CVPR. pp. 14297--14306 (2023)

\bibitem{pan2023slide}
Pan, X., Ye, T., Xia, Z., Song, S., Huang, G.: Slide-transformer: Hierarchical vision transformer with local self-attention. In: CVPR. pp. 2082--2091 (2023)

\bibitem{pan2023effective}
Pan, Z., Gherardi, R., Xie, X., Huang, S.: Effective real image editing with accelerated iterative diffusion inversion. In: ICCV. pp. 15912--15921 (2023)

\bibitem{podell2023sdxl}
Podell, D., English, Z., Lacey, K., Blattmann, A., Dockhorn, T., M{\"u}ller, J., Penna, J., Rombach, R.: Sdxl: Improving latent diffusion models for high-resolution image synthesis. arXiv preprint arXiv:2307.01952  (2023)

\bibitem{radford2021learning}
Radford, A., Kim, J.W., Hallacy, C., Ramesh, A., Goh, G., Agarwal, S., Sastry, G., Askell, A., Mishkin, P., Clark, J., et~al.: Learning transferable visual models from natural language supervision. In: ICML. pp. 8748--8763. PMLR (2021)

\bibitem{rombach2022high}
Rombach, R., Blattmann, A., Lorenz, D., Esser, P., Ommer, B.: High-resolution image synthesis with latent diffusion models. In: CVPR. pp. 10684--10695 (2022)

\bibitem{ronneberger2015u}
Ronneberger, O., Fischer, P., Brox, T.: U-net: Convolutional networks for biomedical image segmentation. In: MICCAI. pp. 234--241. Springer (2015)

\bibitem{russakovsky2015imagenet}
Russakovsky, O., Deng, J., Su, H., Krause, J., Satheesh, S., Ma, S., Huang, Z., Karpathy, A., Khosla, A., Bernstein, M., et~al.: Imagenet large scale visual recognition challenge. IJCV  \textbf{115},  211--252 (2015)

\bibitem{salimans2022progressive}
Salimans, T., Ho, J.: Progressive distillation for fast sampling of diffusion models. arXiv preprint arXiv:2202.00512  (2022)

\bibitem{sauer2023adversarial}
Sauer, A., Lorenz, D., Blattmann, A., Rombach, R.: Adversarial diffusion distillation. arXiv preprint arXiv:2311.17042  (2023)

\bibitem{laion}
Schuhmann, C., Beaumont, R., Vencu, R., Gordon, C., Wightman, R., Cherti, M., Coombes, T., Katta, A., Mullis, C., Wortsman, M., et~al.: Laion-5b: An open large-scale dataset for training next generation image-text models. In: NeurIPS. pp. 25278--25294 (2022)

\bibitem{song2020denoising}
Song, J., Meng, C., Ermon, S.: Denoising diffusion implicit models. In: ICLR (2021)

\bibitem{song2019generative}
Song, Y., Ermon, S.: Generative modeling by estimating gradients of the data distribution. In: NeurIPS. pp. 11895--11907 (2019)

\bibitem{song2020score}
Song, Y., Sohl-Dickstein, J., Kingma, D.P., Kumar, A., Ermon, S., Poole, B.: Score-based generative modeling through stochastic differential equations. In: ICLR (2021)

\bibitem{teng2023relay}
Teng, J., Zheng, W., Ding, M., Hong, W., Wangni, J., Yang, Z., Tang, J.: Relay diffusion: Unifying diffusion process across resolutions for image synthesis. arXiv preprint arXiv:2309.03350  (2023)

\bibitem{xie2023difffit}
Xie, E., Yao, L., Shi, H., Liu, Z., Zhou, D., Liu, Z., Li, J., Li, Z.: Difffit: Unlocking transferability of large diffusion models via simple parameter-efficient fine-tuning. arXiv preprint arXiv:2304.06648  (2023)

\bibitem{yang2023diffusion}
Yang, X., Zhou, D., Feng, J., Wang, X.: Diffusion probabilistic model made slim. In: CVPR. pp. 22552--22562 (2023)

\bibitem{zheng2023any}
Zheng, Q., Guo, Y., Deng, J., Han, J., Li, Y., Xu, S., Xu, H.: Any-size-diffusion: Toward efficient text-driven synthesis for any-size hd images. arXiv preprint arXiv:2308.16582  (2023)

\end{thebibliography}
\clearpage 
\title{Appendix for HiDiffusion:  Unlocking Higher-Resolution Creativity and Efficiency in Pretrained Diffusion Models} 

\titlerunning{HiDiffusion}


\authorrunning{Shen Zhang et al.}


\author{\vspace{-0.5cm}}

\institute{\vspace{-0.5cm}}

\maketitle
\renewcommand\thesection{\Alph{section}}

\begin{figure}[htbp]
  \centering
  \includegraphics[width=\textwidth]{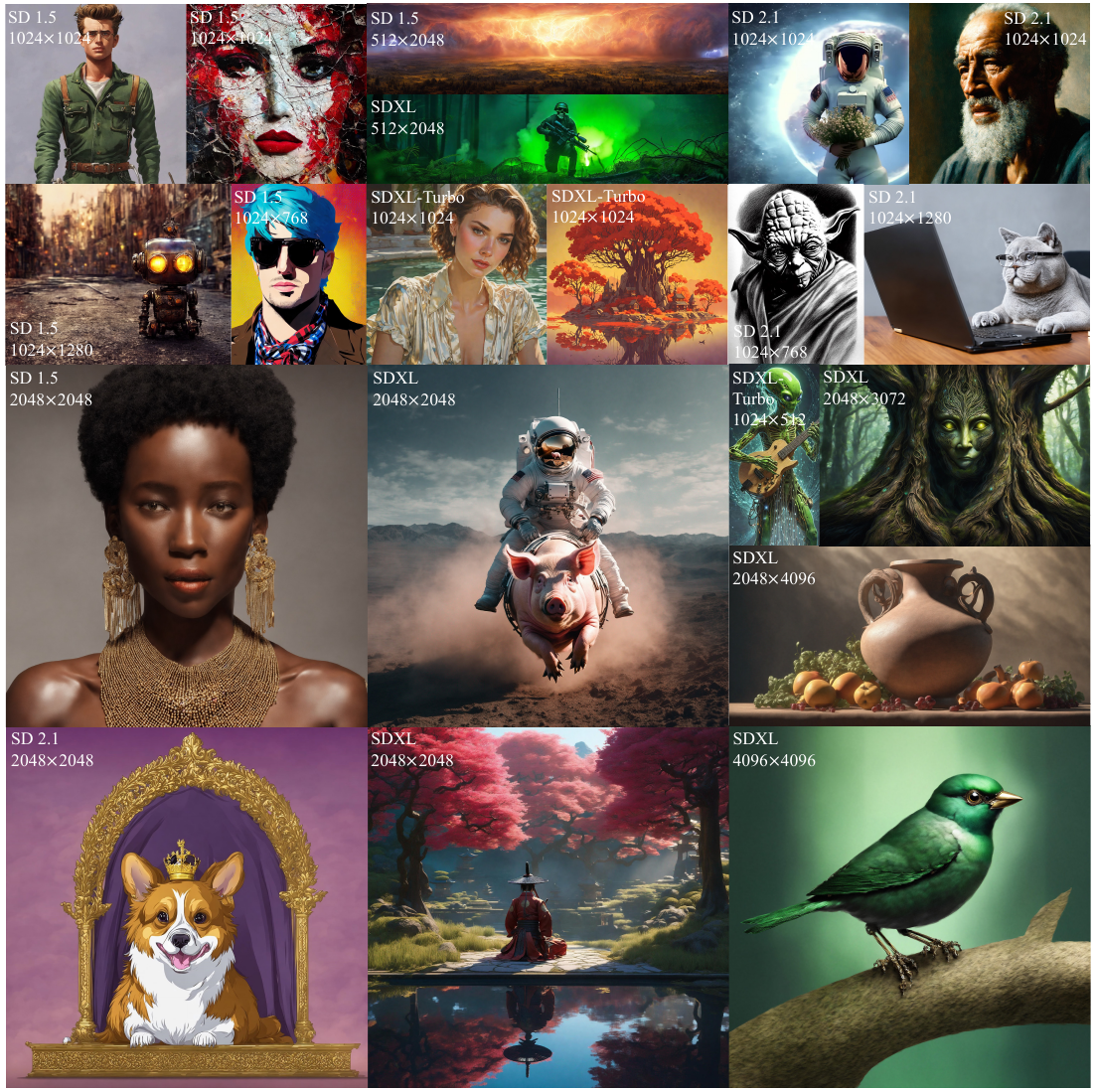}
      \caption{Select HiDiffusion samples for various diffusion models, resolutions, and aspect ratios.  HiDiffusion enables pretrained diffusion models to generate higher-resolution images surpassing the training image size without further training or fine-tuning and can effectively accelerate the inference. Best viewed when zoomed in.
  }
  \label{fig:imgae_gallery}
  \vspace{-2em}
\end{figure}

\clearpage

In the appendix, we present the following details associated with HiDiffusion:
\begin{itemize}
[noitemsep,nolistsep,topsep=0pt,parsep=0pt,partopsep=0pt]

\item  Visualization of feature duplication across inference steps.
\item More ablations about the components of HiDiffusion, including the effect of RAU-Net and MSW-MSA, the RAU operation, the position of RAD and RAU, the Switching Threshold, the window size of MSW-MSA.

\item Details about SD 2.1, SDXL, SDXL-Turbo settings.
\item Details about extreme resolutions~(2048$\times$2048 for SD 1.5, SD 2.1, 4096$\times$4096 for SDXL).
\item Extensions to image-to-image task.
\item More visualization results, including comparisons to diffusion acceleration and high-resolution synthesis methods.

\end{itemize}

\section{Feature Duplication across Inference Step}
\begin{figure}[htbp]
  \centering
  \includegraphics[width=0.8\textwidth]{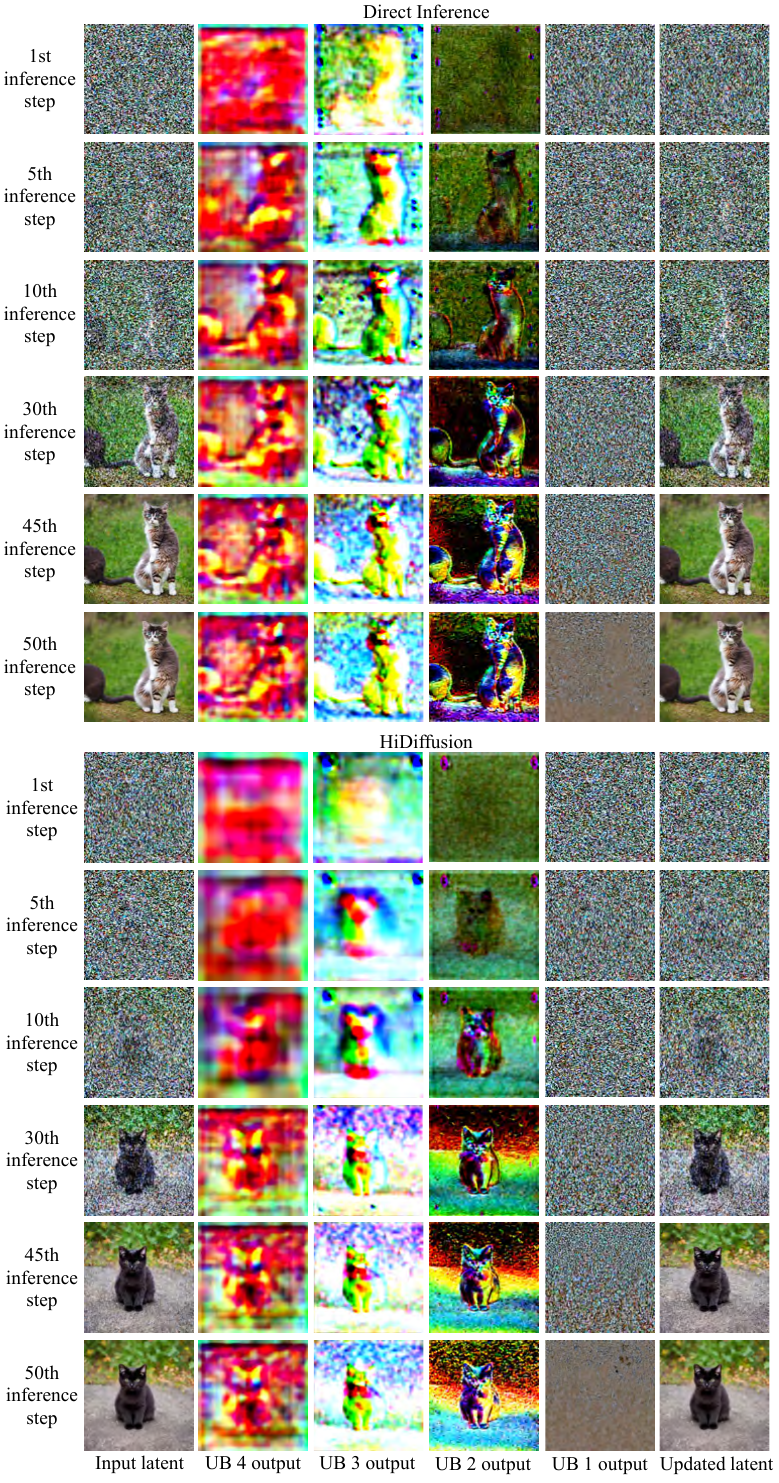}
      \caption{The feature map visualization across different inference steps based on SD 1.5. The image resolution is 1024$\times$1024 and we adopt 50 DDIM steps.
  }
  \label{fig:feature_timestep}
  \vspace{-2em}
\end{figure}

When directly inferring to generate higher-resolution images using pretrained diffusion models, we observed the feature duplication phenomenon at the 30th inference step. This section presents feature visualization across different inference steps to demonstrate that feature duplication arises at almost every inference step, as shown in~\cref{fig:feature_timestep}. Even when the input latent is extremely noisy in the early denoising stages, such as the 1st inference step, direct inference still leads to conspicuous feature duplication, as shown in the UB 3 output of the 1st inference step.
When the input latent is less noisy, for instance, at the 45th inference step, more severely pronounced feature duplication emerges. The feature duplication impacts the trajectory of image generation, ultimately causing object duplication in the final output image~(the updated latent at the 50th inference step). Compared to direct inference, our HiDiffusion effectively alleviates feature duplication at each inference step and can generate reasonable higher-resolution images in a tuning-free way.

\section{More Ablation Results}

\begin{figure}[htbp]
  \centering
  \includegraphics[width=\textwidth]{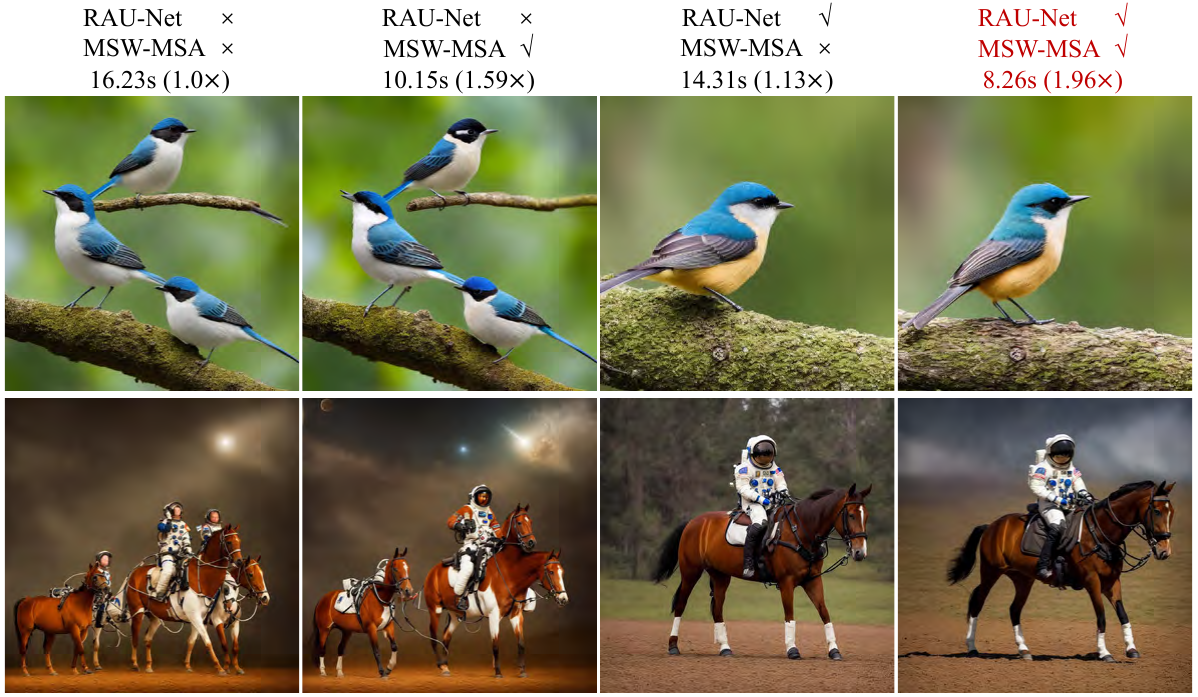}
      \caption{The effect of RAU-Net and MSW-MSA based on SD 1.5. The resolution is 1024$\times$1024. 
  }
  \label{fig:raunet_mswmsa}
  \vspace{-2em}
\end{figure}

\subsection{The Effect of RAU-Net and MSW-MSA}
We have analyzed the impact of RAU-Net and MSW-MSA in the main paper. Here we provide the qualitative comparison of all possible combinations, as shown in~\cref{fig:raunet_mswmsa}.

\subsection{The RAD Operation}
In the main paper, RAD is achieved by altering the stride, padding, and dilation of the original downsampler's convolution.  Alternatively,  We can add an extra adaptive pooling and keep the convolution unchanged, which can be written as:
\begin{equation}
\mathcal{R}(\mathcal{C}_{3,1,2,1}(x), \alpha)=ada\_pool(\mathcal{C}_{3,1,2,1}(x), \frac{\alpha}{2}).
\end{equation}
This method can also achieve the goal of resolution-aware downsampling. for the 1024$\times$1024 resolution generation, $\alpha$ is set as 4. In this section, we investigate which methods can generate higher-quality images. We present quantitative comparison in~\cref{table:rao_var}. Compared to the additional pooling operation, the method used in the main paper exhibits better performance in both FID and CLIP-Score.

\begin{table}[htbp]
    \centering \vspace{-0.5em}
    \resizebox{0.43\linewidth}{!}{%
    \begin{tabular}{l c c }
        \toprule
         Method     &    FID$\downarrow$              &  CLIP-Score$\uparrow$ \\
         \midrule
         Adaptive pooling         &  26.73   &   0.304        \\
        Larger stride    &     21.81    &    0.307    \\
        \bottomrule
    \end{tabular}
    }
     \caption{Quantitative evaluation of two variants of resolution-aware operation in zero-shot text-guided image synthesis on ImageNet based on SD 1.5. The resolution is 1024$\times$1024.
        }
    \label{table:rao_var}
\end{table}

\subsection{The impact of the position of RAD and RAU}

\begin{figure}[htbp]
  \centering
  \includegraphics[width=0.7\textwidth]{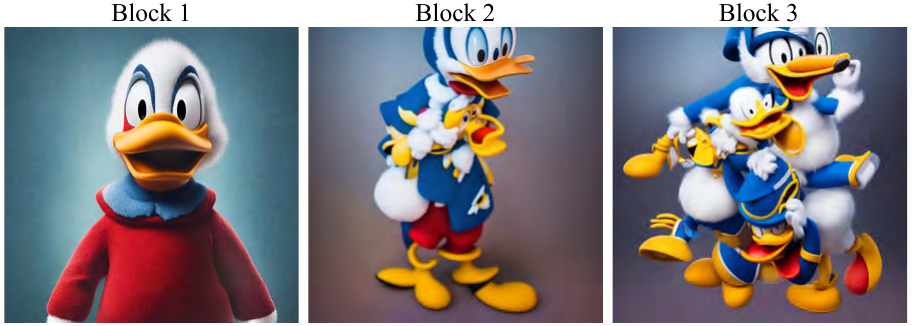}
      \caption{Qualitative comparison between different positions of RAD and RAU based on SD 1.5. The resolution is 1024$\times$1024. 
  }
  \label{fig:diff_block}
\end{figure}

\begin{table}[htbp]
    \centering\vspace{-0.7em}
    \resizebox{0.45\linewidth}{!}{%
    \begin{tabular}{l c c c }
        \toprule
         Position      & Block 1                 & Block 2 & Block 3  \\
         \midrule
         FID          &  21.81  &    20.84       &    21.26         \\
        CLIP-Score    &    0.307                        &   0.307          &    0.305           \\
        \bottomrule
        \vspace{-1.2em}
    \end{tabular}
    }
     \caption{Quantitative evaluation of the position of RAD and RAU in zero-shot text-guided image synthesis on ImageNet based on SD 1.5. The resolution is 1024$\times$1024.
        \vspace{-1em}}
    \label{table:RAD_RAU_impact}
\end{table}
Our main idea is to introduce RAD and RAU to dynamically downsample the feature map. We insert the RAD and RAU into Block 1, Block 2, and Block 3 respectively to examine the impact of the Resolution-aware sampler at different locations, as shown in \cref{table:RAD_RAU_impact}. There is a minor quantitative metric difference between different locations. However, we visually observe that incorporating  RAD and RAU in Block 1 can better mitigate object duplication, as illustrated in~\cref{fig:diff_block}.

\subsection{The Switching Threshold}
\begin{figure}[htbp]
  \centering
  \includegraphics[width=\textwidth]{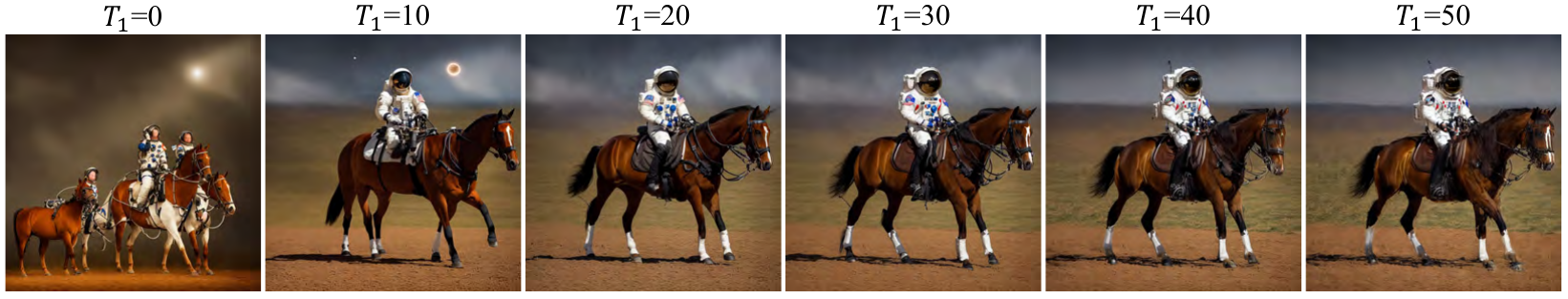}
      \caption{Qualitative comparison between different $T_1$ based on SD 1.5. The resolution is 1024$\times$1024. 
  }
  \label{fig:diff_t1}
  \vspace{-2em}
\end{figure}

\begin{table}[!h]
    \centering \vspace{-1em}
    \resizebox{0.53\linewidth}{!}{%
    \begin{tabular}{l c c c c c c}
        \toprule
         $T_1$      & 0                 & 10 & 20 & 30&40&50 \\
         \midrule
         FID          & 26.38    &  24.05          &    21.81 &    21.18  & 21.12 &25.55        \\
        CLIP-Score    &   0.309                        &    0.308        & 0.307   & 0.308  & 0.307  &   0.295       \\
        \bottomrule
    \end{tabular}
    }
     \caption{Quantitative evaluation of different $T_1$ in zero-shot text-guided image synthesis on ImageNet based on SD 1.5. The resolution is 1024$\times$1024.
        \vspace{-1em}}
    \label{table:abl_T}
\end{table}

\begin{figure}[]
  \centering
  \includegraphics[width=\textwidth]{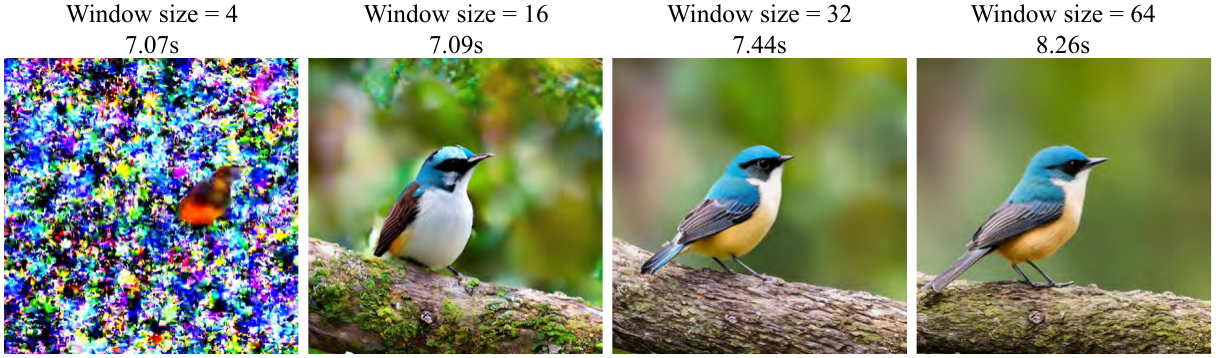}
      \caption{Qualitative comparison between different window sizes based on SD 1.5. The resolution is 1024$\times$1024. 
  }
  \label{fig:window_size}
\end{figure}

\begin{table}[]
    \centering
    \resizebox{0.4\linewidth}{!}{%
    \begin{tabular}{l c c c c}
        \toprule
         Window size      & 4                  & 16 & 32&64 \\
         \midrule
         Latency~(s)          &  7.07  &  7.09           &  7.44   &  8.26          \\
         FID          &  417.15  &  53.02           &  22.37   &  21.81          \\
        CLIP-Score    & 0.225                           &   0.295          &   0.307  &    0.307        \\
        \bottomrule
    \end{tabular}
    }
     \caption{Quantitative evaluation of the window size in zero-shot text-guided image synthesis on ImageNet based on SD 1.5. The resolution is 1024$\times$1024.
        }
    \label{table:window_size}
\end{table}

The Switching Threshold determines when to switch from RAU-Net to vanilla U-Net. We explore the impact of different thresholds on the performance of HiDiffusion.  The quantitative results are shown in \cref{table:abl_T}. $T_1$=0 indicates that RAU-Net is not utilized, while $T_1$=50 indicates that RAU-Net is used in the entire denoising process. When $T_1$ is between 20 and 40, we achieve the best performance in metric evaluation. The qualitative comparison demonstrates that $T_1$ ranging from 20 to 40 can effectively alleviate object duplication, with $T_1=20$ yielding the optimal performance, as shown in~\cref{fig:diff_t1}.
Therefore, we select $T_1=20$ as the default setting.

\subsection{The Window Size of MSW-MSA}

The window size determines the receptive field of self-attention. We compare the performance from the small window size proposed in Swin Transformer to our proposed large window size based on SD 1.5, as shown in \cref{table:window_size}. As the window size gradually increases, the performance improves. We achieve the optimal balance between efficiency and image quality when the window size is half the height and width of the feature map. The qualitative results are shown in~\cref{fig:window_size}.

\section{Details about Other Models settings.}
\begin{figure}[!h]
     \centering
     \begin{subfigure}[b]{0.3\textwidth}
         \centering
         \includegraphics[width=\textwidth]{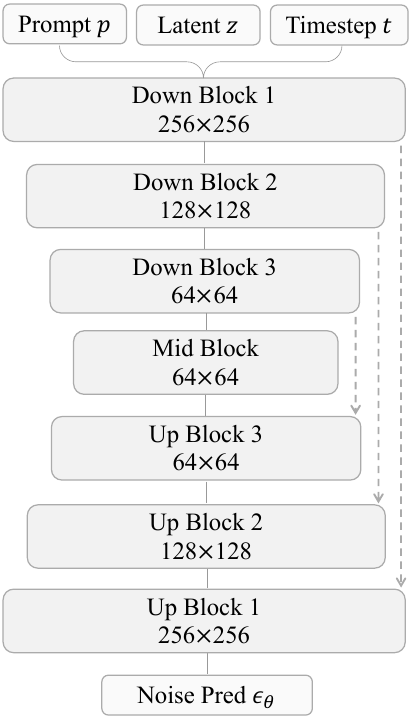}
         \caption{Vanilla U-Net.} \label{fig:vanilla_unet_sdxl}
     \end{subfigure}
     \begin{subfigure}[b]{0.302\textwidth}
         \centering
         \includegraphics[width=\textwidth]{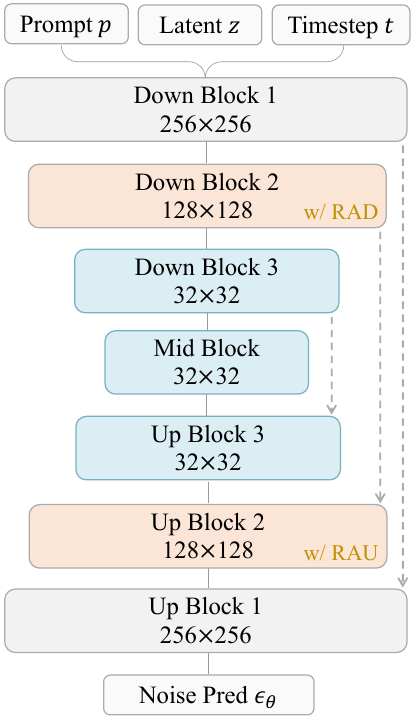}
         \caption{HiDiffusion RAU-Net.}
     \end{subfigure}
    \caption{Comparison between vanilla SDXL’s U-Net and our proposed HiDiffusion RAU-Net for SDXL. Parameters in all blocks are frozen. The main difference lies in the \textcolor{cyan}{blue} Blocks (differ in the dimensions of feature map) and \textcolor{orange}{orange} Blocks (Our proposed RAD and RAU modules are incorporated into \textcolor{orange}{Block 2}). } \label{fig:model_architecture_sdxl}
    \label{fig:sdxl_unet}
    \vspace{-10pt}
\end{figure}

\begin{figure}[!h]
     \centering
     \begin{subfigure}[b]{0.3\textwidth}
         \centering
         \includegraphics[width=\textwidth]{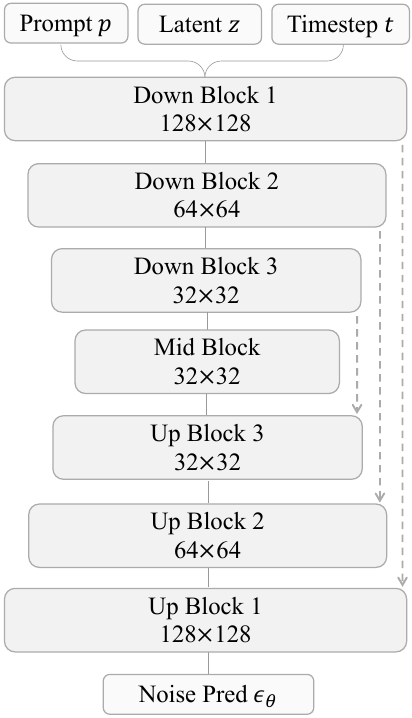}
         \caption{Vanilla U-Net.} \label{fig:vanilla_unet_sdxl_turbo}
     \end{subfigure}
     \begin{subfigure}[b]{0.301\textwidth}
         \centering
         \includegraphics[width=\textwidth]{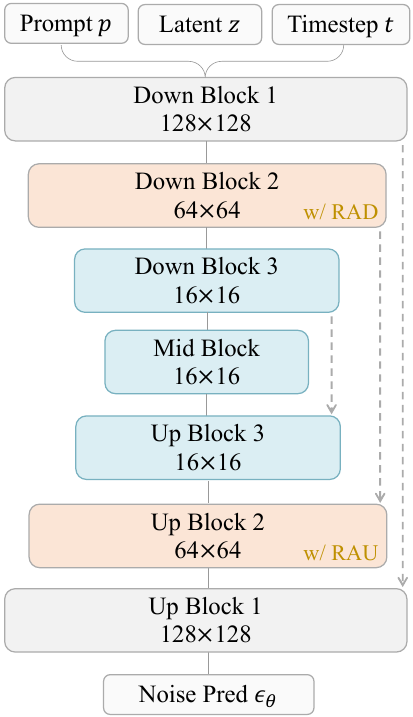}
         \caption{HiDiffusion RAU-Net.}
     \end{subfigure}
    \caption{Comparison between vanilla SDXL-Turbo’s U-Net and our proposed HiDiffusion RAU-Net for SDXL-Turbo. Parameters in all blocks are frozen. The main difference lies in the \textcolor{cyan}{blue} Blocks (differ in the dimensions of feature map) and \textcolor{orange}{orange} Blocks (Our proposed RAD and RAU modules are incorporated into \textcolor{orange}{Block 2}.). } \label{fig:model_architecture_sdxl_turbo}
    \label{fig:sdxl_turbo_unet}
    \vspace{-10pt}
\end{figure}

\begin{figure} 
	\centering 
	\includegraphics[width=1.\linewidth]{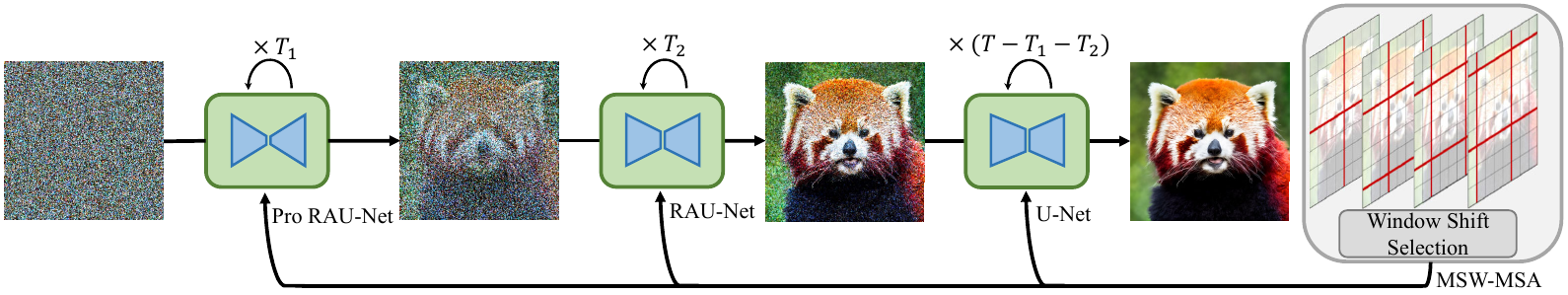}
	\caption{The framework of image synthesis with extreme resolution~(2048$\times$2048 for SD 1.5 and SD 2.1, 4096$\times$4096 for SDXL). Pro RAU-Net: progressive RAU-Net.} 
	\label{fig:2048_framework}
\end{figure}

\begin{figure}[!h] 
	\centering 
	\includegraphics[width=0.5\linewidth]{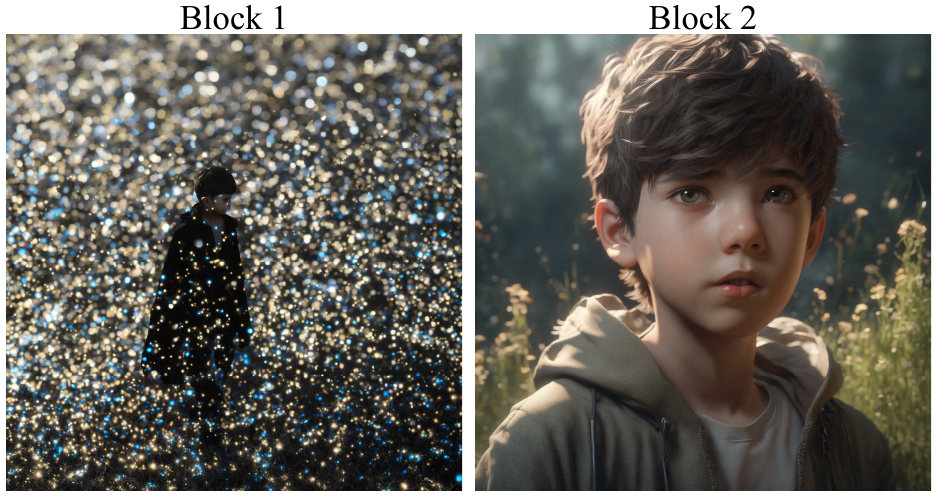}
	\caption{2048$\times$2048 resolution comparison between inserting resolution-aware samplers into Block 1 and Block 2 based on SDXL.} 
	\label{fig:block1_block2}
	\vspace{-0.6cm}
\end{figure}

As SD 1.5 and SD 2.1 share the same U-Net architecture, the settings of SD 2.1 are consistent with those of SD 1.5.

An illustrative comparison of the vanilla SDXL U-Net and RAU-Net for SDXL in the context of generating 2048$\times$2048 resolution images is presented in~\cref{fig:sdxl_unet}. We incorporate RAD and RAU in Block 2 and set $\alpha=\beta=4$ to match the deep blocks of U-Net. In contrast to SD 1.5 and SD 2.1's U-Net, Down Block 1 and Up Block 1 of SDXL only consist of two and three ResNet blocks, respectively. If we choose to incorporate the RAD and RAU in Block 1, the ResNet Blocks in Block 1 are insufficient to effectively handle the resolution change caused by the interpolation function in RAD, resulting in the synthesis of blurry images. We present the qualitative comparison between inserting RAD and RAU in Block 1 and inserting RAD and RAU in Block 2 in~\cref{fig:block1_block2}. In the experiment of the main paper, We set $T_1=20$ for 50 DDIM steps. The classifier-free guidance scale is 7.5. Since Block 1 of SDXL U-Net does not contain self-attention, we incorporate MSW-MSA into Block 2. We set the window size as $(64, 64)$. The predefined set of shift strides is $\{(0,0), (16,16),(32,32),(48,48)\}$. For 4096$\times$4096 resolution generation, please refer to~\cref{extreme_res}.

The U-Net architecture of SDXL-Turbo and SDXL are very similar, except for the differences in input and output dimensions, as shown in~\cref{fig:sdxl_turbo_unet}. We introduce the setting of SDXL-Turbo in brief. We incorporate the RAD and RAU into Block 2. The inference step is 4 and we set $T_1=2$. Classifier-free guidance is not used. We set the window size as $(32, 32)$. The predefined set of shift strides is $\{(0,0), (8,8),(16,16),(24,24)\}$.

\section{Details about extreme resolutions}
\label{extreme_res}
\begin{figure}
     \centering
     \begin{subfigure}[b]{0.302\textwidth}
         \centering
         \includegraphics[width=\textwidth]{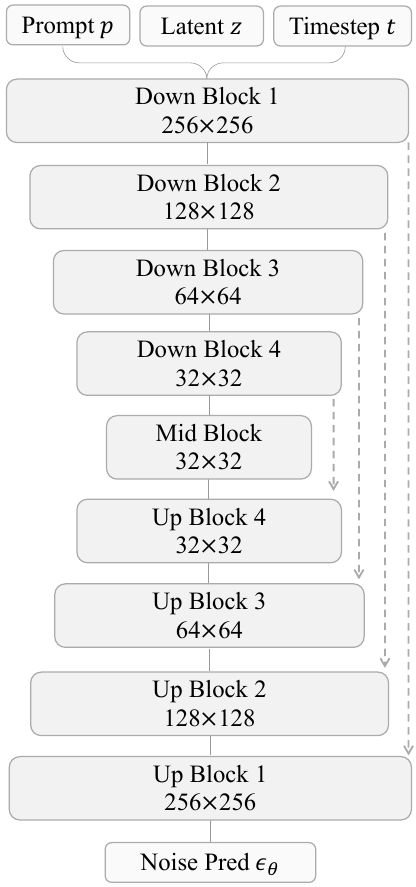}
         \caption{} \label{fig:vanilla_unet_2048}
     \end{subfigure}
     \begin{subfigure}[b]{0.3\textwidth}
         \centering
         \includegraphics[width=\textwidth]{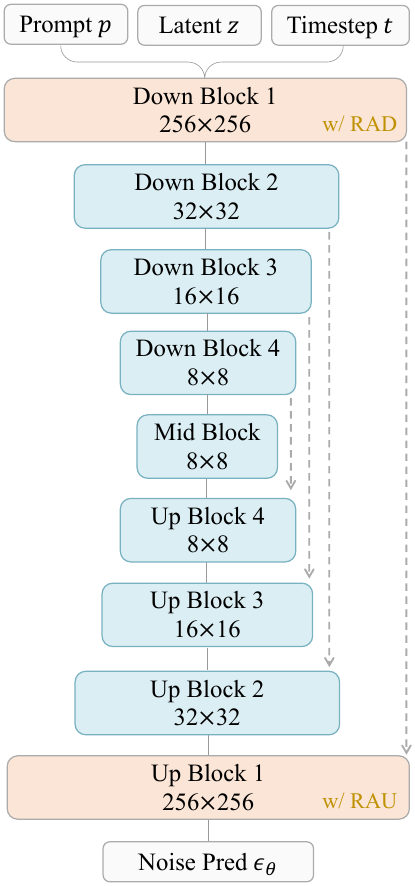}
         \caption{} \label{fig:rau-net_2048}
     \end{subfigure}
     \begin{subfigure}[b]{0.3\textwidth}
         \centering
         \includegraphics[width=\textwidth]{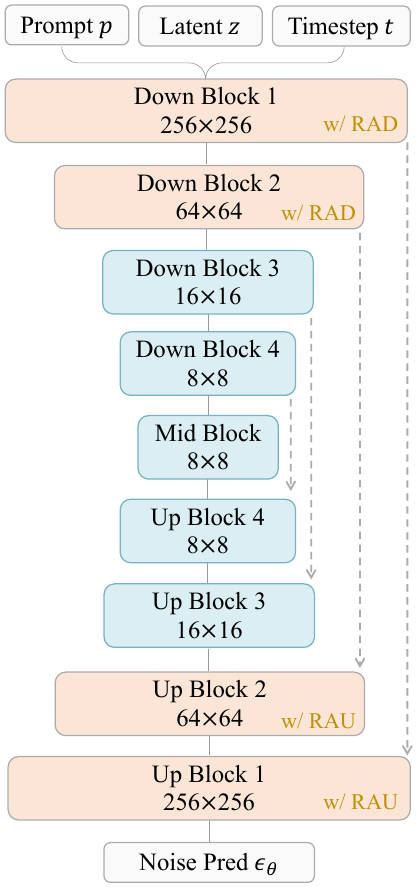}
         \caption{} \label{fig:rau-net_progressive_2048}
     \end{subfigure}
    \caption{U-Net variants of SD 1.5 and SD 2.1. (a) Vanilla U-Net. (b) RAU-Net. (c) Progressive RAU-Net.}
    \label{fig:extreme_2048}
\end{figure}

\begin{figure}
     \centering
     \begin{subfigure}[b]{0.3\textwidth}
         \centering
         \includegraphics[width=\textwidth]{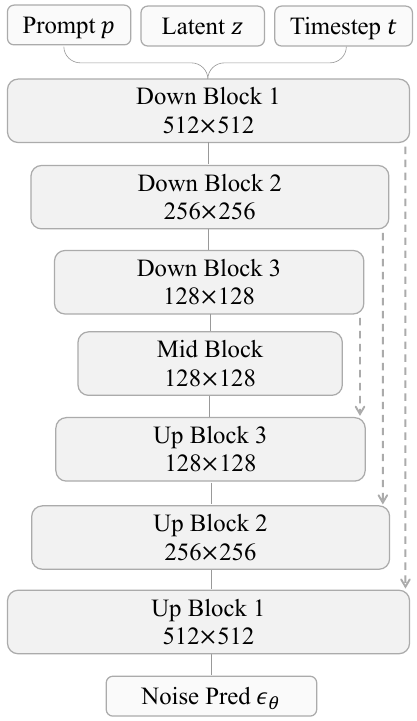}
         \caption{} \label{fig:vanilla_unet_2048_sdxl}
     \end{subfigure}
     \begin{subfigure}[b]{0.3\textwidth}
         \centering
         \includegraphics[width=\textwidth]{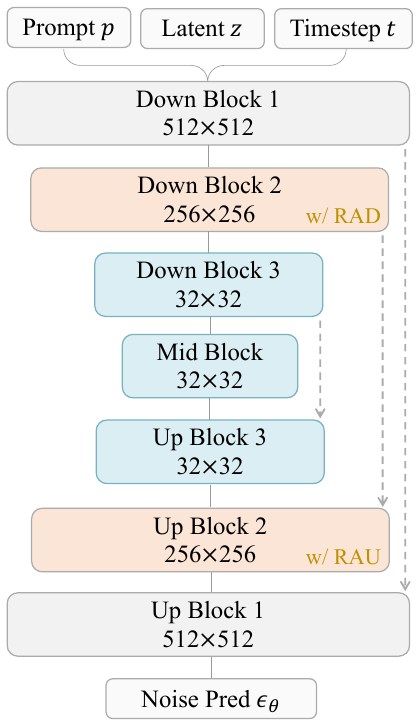}
         \caption{} \label{fig:rau-net_2048_sdxl}
     \end{subfigure}
     \begin{subfigure}[b]{0.3\textwidth}
         \centering
         \includegraphics[width=\textwidth]{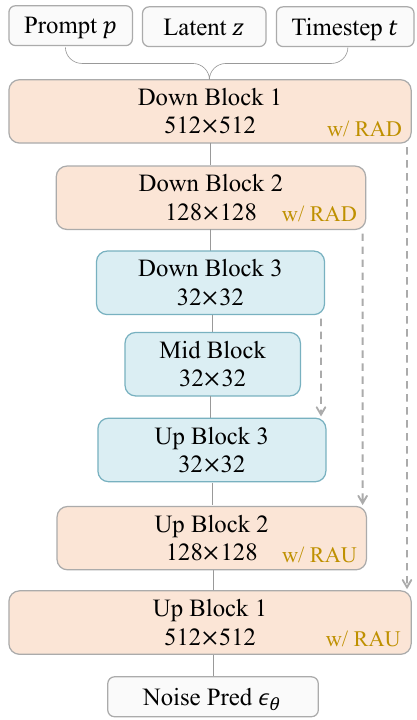}
         \caption{} \label{fig:rau-net_progressive_2048_sdxl}
     \end{subfigure}
    \caption{U-Net variants of SDXL. (a) Vanilla U-Net. (b) RAU-Net. (c) Progressive RAU-Net. The parameter settings of (c) are same with~\cref{fig:extreme_2048}.}
    \label{fig:extreme_4096}
\end{figure}

\begin{figure}
     \centering
     \begin{subfigure}[b]{0.3\textwidth}
         \centering
         \includegraphics[width=\textwidth]{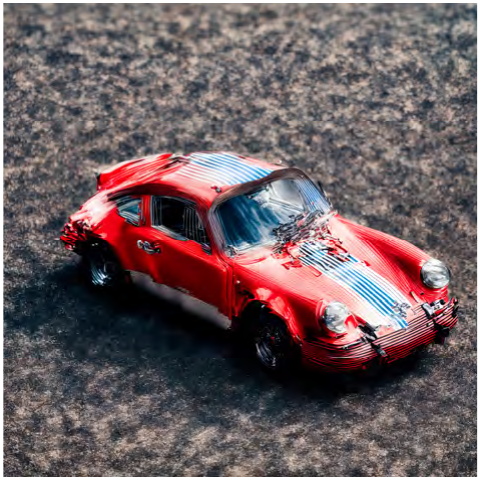}
         \caption{} \label{fig:2048_1}
     \end{subfigure}
     \begin{subfigure}[b]{0.3\textwidth}
         \centering
         \includegraphics[width=\textwidth]{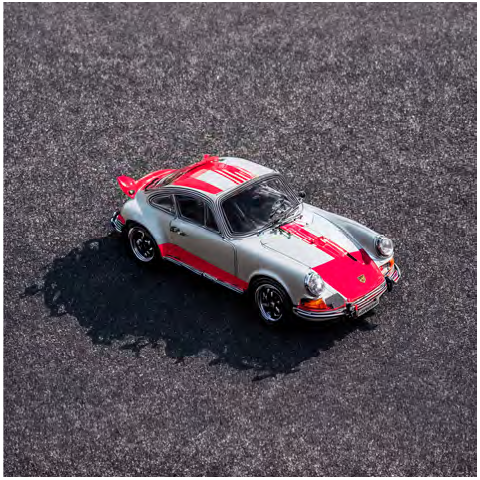}
         \caption{} \label{fig:2048_2}
     \end{subfigure}
    \caption{2048$\times$2048 resolution samples generated by (a) Directly set $\alpha=8$ in Block 1 of RAU-Net. (b) The final progressive method. The diffusion model version is SD 1.5.}
    \label{fig:extreme_2048_sample}
    \vspace{-10pt}
\end{figure}

\begin{figure}[!h] 
	\centering 
	\includegraphics[width=0.7\linewidth]{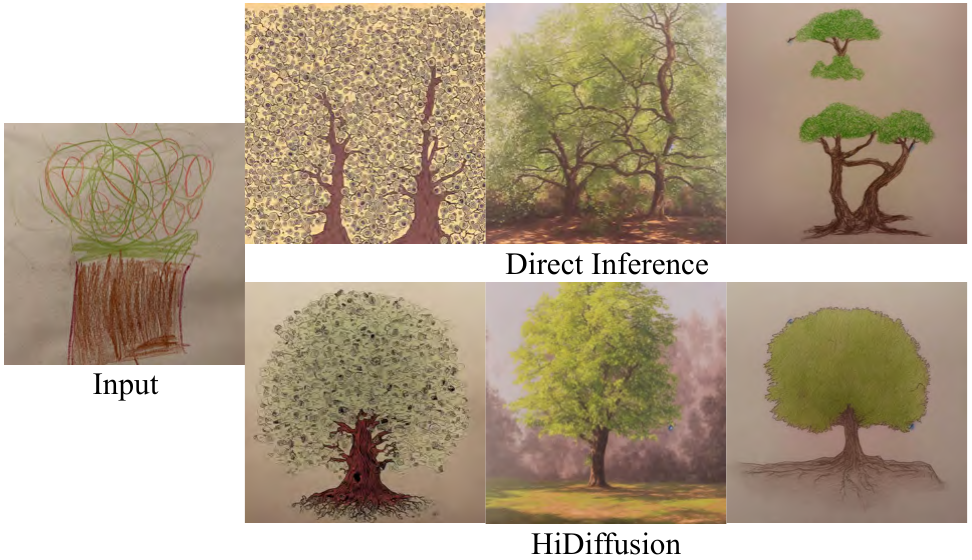}
	\caption{SDEdit task of 1024$\times$1024 resolution based on SD 1.5.} 
	\label{fig:img2img_sdedit}
\end{figure}

\begin{figure}[!h] 
	\centering 
	\includegraphics[width=0.7\linewidth]{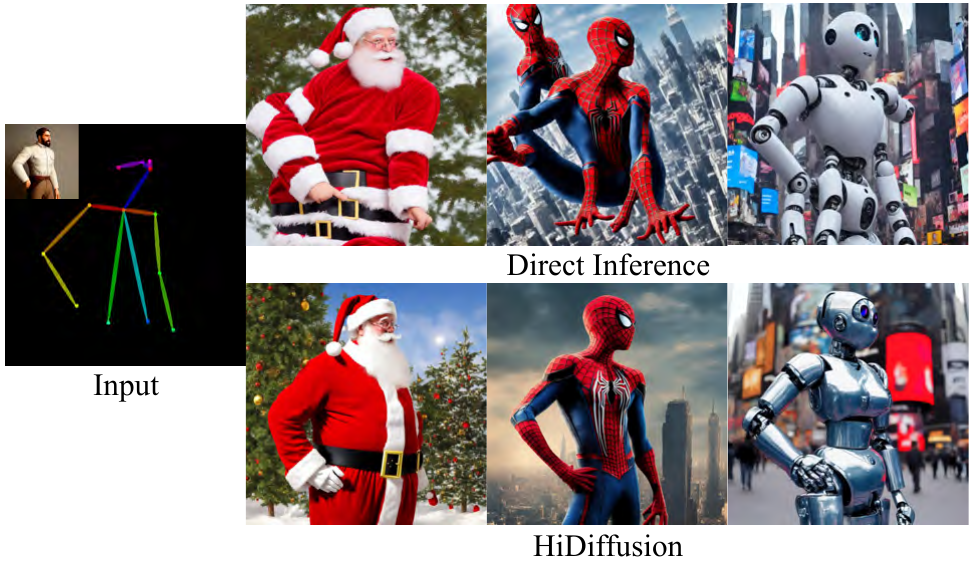}
	\caption{ControlNet task of 1024$\times$1024 resolution based on SD 1.5.} 
	\label{fig:img2img_controlnet}
\end{figure}

\begin{figure}[htbp]
  \centering
  \includegraphics[width=\textwidth]{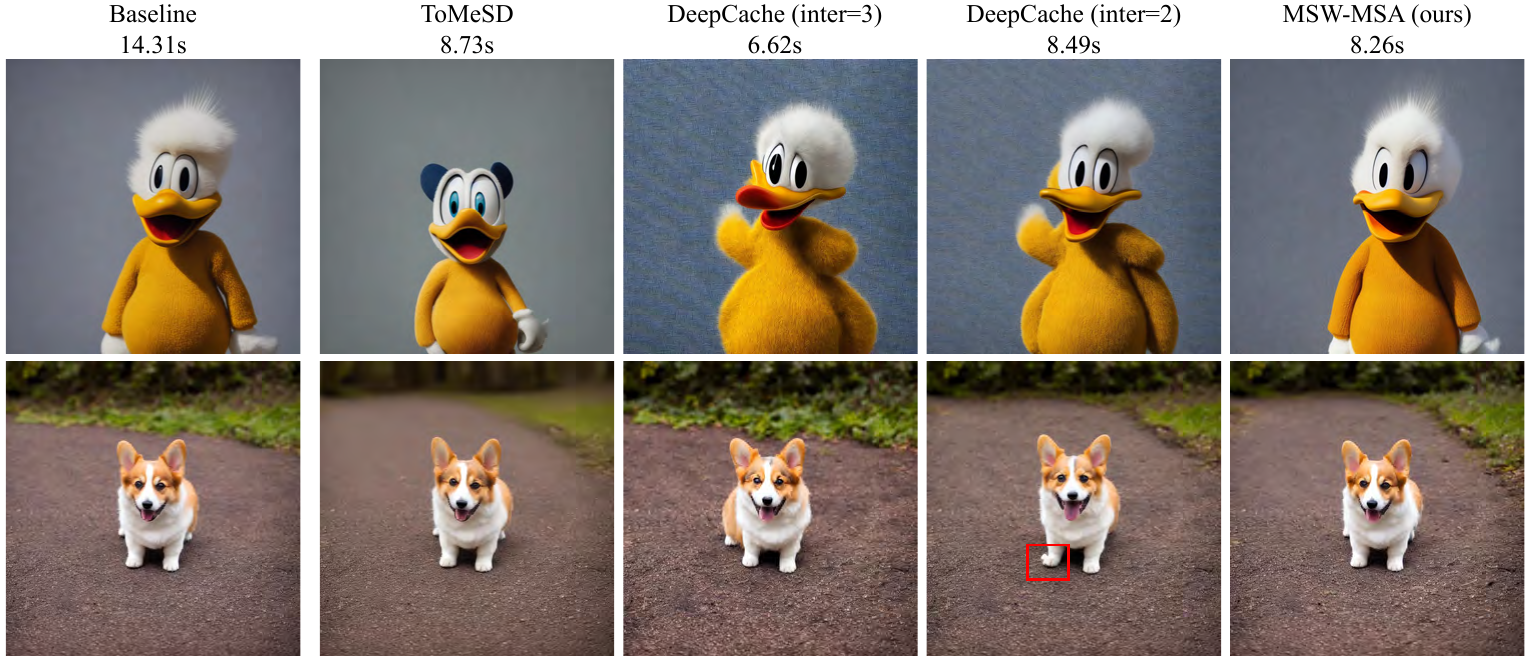}
      \caption{The qualitative comparison between different diffusion acceleration methods based on SD 1.5. The resolution is 1024$\times$1024. The baseline is SD 1.5 with RAU-Net.
  }
  \label{fig:deepcache_compare}
  \vspace{-2em}
\end{figure}

For SD 1.5 and SD 2.1, generating images with 2048$\times$2048 resolution is a significant challenge, considering that this resolution is already 16 times the training image resolution. RAU-Net can generate images with 2048$\times$2048 resolution by simply setting $\alpha=\beta=8$, as shown in~\cref{fig:rau-net_2048}.  However, $\beta=8$ implies that RAU  upsamples the feature map by a factor of 8 using an interpolation function. This abrupt resolution change brought by interpolation leads to the generation of blurry images, as illustrated in~\cref{fig:2048_1}.  To tackle the issue, we adopt a progressive variant of RAU-Net, as shown in~\cref{fig:rau-net_progressive_2048}. We incorporate RAU and RAD with $\alpha=\beta=4$ into Block 1 and Block 2, respectively. 
This allows the feature map to gradually align with the deep blocks of U-Net, thus circumventing the blurriness issue caused by a large interpolation factor. For 4096$\times$4096 resolution generation of SDXL,  we also adopt progressive RAU-Net, as shown in~\cref{fig:rau-net_progressive_2048_sdxl}. We incorporate RAU and RAD with $\alpha=\beta=4$ into Block 1 and Block 2, respectively.

As described in the main paper, matching the feature map size with the deep blocks of U-Net can generate coherent object structures while potentially affecting image details. Therefore, when generating images with extreme resolution, we gradually reduce the usage of resolution-aware samplers throughout the denoising process for finer image detail. Specifically, we employ Progressive RAU-Net in the early stage, followed by RAU-Net in the middle stage, and finally vanilla U-Net in the later stage. We establish two thresholds $T_1$ and $T_2$: when denoising steps $t < T_1$, We use progressive RAU-Net;
when $T_1 \leq t \leq T_2$, We use RAU-Net
; when $t > T_2$, vanilla U-Net is used. We present the framework in~\cref{fig:2048_framework} and generated samples in~\cref{fig:2048_2}. In the experiment of the main paper, We set $T_1=15$ and $T_2=35$ for 50 DDIM steps.  We incorporate MSW-MSA into Block 1 for SD 1.5 and SD 2.1, and into Block 2 for SDXL. We set the window size as $(128, 128)$. The predefined set of shift strides is $\{(0,0), (32,32),(64,64),(96,96)\}$.  The classifier-free guidance scales of SD 1.5, SD 2.1, and SDXL are all 7.5.

\section{Image-to-image Task}
HiDiffusion also works well in image-to-image tasks including SDEdit and ControlNet, as shown in~\cref{fig:img2img_sdedit,fig:img2img_controlnet}.

\section{More Visualization Results}
\subsection{Comparison to Diffusion Acceleration Methods}
In the main paper, We have demonstrated the effectiveness of our MSW-MSA compared to other diffusion acceleration methods. Here we provide the qualitative comparison between MSW-MSA and other methods, as shown in~\cref{fig:deepcache_compare}. Compared to other methods, the generated images by our method are more consistent with the baseline. Furthermore, we surpass other methods in terms of details and local features.

\subsection{Comparison to High-Resolution Synthesis Method}
\begin{figure}[htbp]
  \centering
  \includegraphics[width=\textwidth]{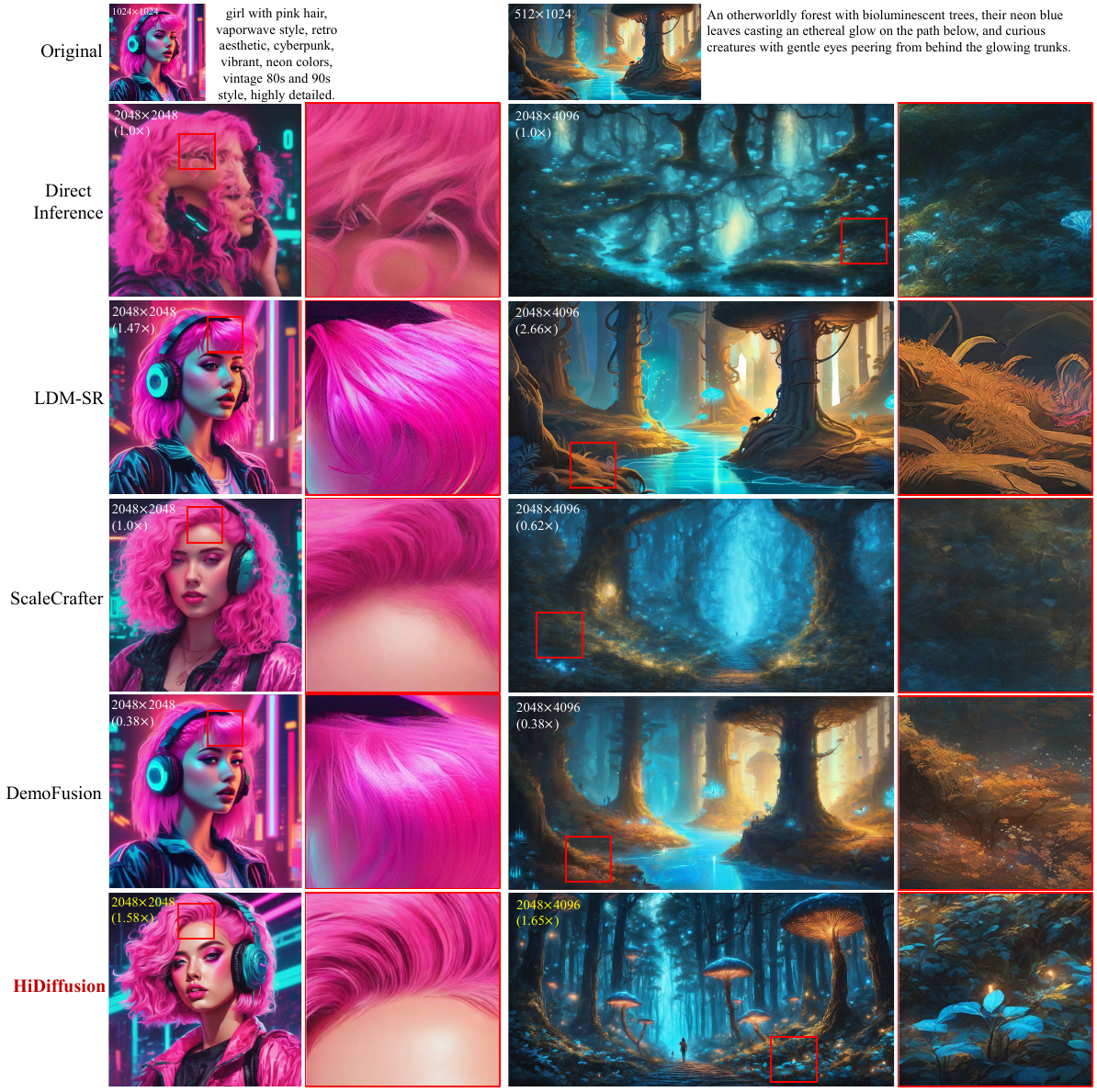}
      \caption{More qualitative comparison with other methods based on SDXL. The input prompt is located to the right of the original image. The first line of text in the image indicates the image resolution, while the second line indicates the inference speed relative to direct inference. Best viewed when zoomed in.
  }
  \label{fig:image_compare_high1}
  \vspace{-2em}
\end{figure}

\begin{figure}[htbp]
  \centering
  \includegraphics[width=\textwidth]{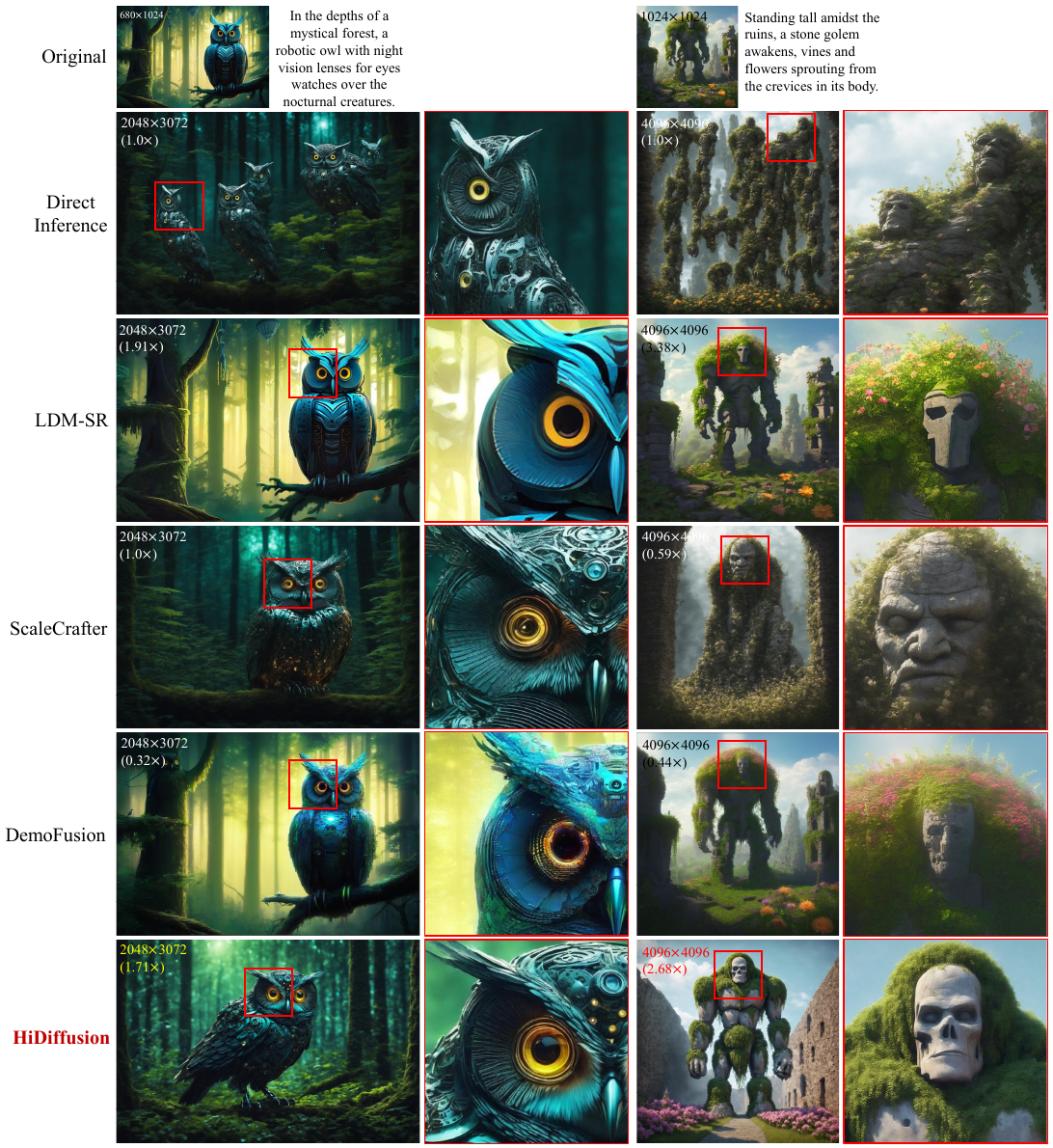}
      \caption{More qualitative comparison with other methods based on SDXL. The input prompt is located to the right of the original image. The first line of text in the image indicates the image resolution, while the second line indicates the inference speed relative to direct inference. Best viewed when zoomed in.
  }
  \label{fig:image_compare_high2}
  \vspace{-2em}
\end{figure}

We present more comparison results with LDM-SR, ScaleCrafter, and Demofusion based on SDXL to demonstrate the effectiveness of HiDiffusion, as shown in~\cref{fig:image_compare_high1,fig:image_compare_high2}. HiDiffusion outperforms ScaleCrafter and DemoFusion in local details, while compared to LDM-SR, HiDiffusion achieves comparable or even better performance.

\end{document}